\documentclass[]{fairmeta}
\usepackage{xcolor}         % colors
\definecolor{RowColor}{rgb}{0.93, 0.98, 0.97}
\definecolor{deepgray}{rgb}{0.5, 0.5, 0.5}
\definecolor{Salmon}{rgb}{1.0, 0.55, 0.41}
\definecolor{RoyalBlue}{rgb}{0.25, 0.41, 1.0}
\usepackage[utf8]{inputenc} % allow utf-8 input
\usepackage{url}            % simple URL 
\usepackage{booktabs}       % professional-quality tables
\usepackage{hhline}
\usepackage{amsbsy,amsmath}
\usepackage{nicefrac}       % compact symbols for 1/2, etc.
\usepackage{microtype}      % microtypography
\usepackage{colortbl}
\usepackage{graphicx}
\usepackage{amsmath}
\usepackage{amssymb}
\usepackage{marvosym}
\usepackage{booktabs}
\usepackage{multirow}
\usepackage{makecell}
\usepackage{url}            % simple URL typesetting
\usepackage{booktabs}       % professional-quality tables
\usepackage{amsfonts}       % blackboard math symbols
\usepackage{nicefrac}       % compact symbols for 1/2, etc.
\usepackage{microtype}      % microtypography
\usepackage{xcolor}         % colors
\usepackage{enumitem}
\usepackage{graphicx}
\usepackage{multirow}
\usepackage{amsmath}
\usepackage{amstext}
\usepackage{MnSymbol}
\usepackage{wasysym}
\usepackage{listings}
\usepackage{color,colortbl}
\usepackage{wrapfig}
\usepackage{soul}
\usepackage{arydshln}
\usepackage{bm}
\usepackage{algorithm}
\usepackage{algpseudocode}
\usepackage{textcomp}
\usepackage{multirow}
\usepackage{subcaption}
\usepackage{tabularx}
\usepackage{adjustbox}
\usepackage{makecell}
\usepackage{array}
\usepackage{amssymb}% http://ctan.org/pkg/amssymb
\usepackage{graphicx}

\title{
Token-Shuffle: Towards High-Resolution Image Generation with Autoregressive Models
}

\author[1*]{Xu Ma}
\author[3]{Peize Sun}
\author[2]{Haoyu Ma}
\author[3]{Hao Tang}
\author[2]{Chih-Yao Ma}
\author[2]{Jialiang Wang}
\author[2]{Kunpeng Li}
\author[2]{Xiaoliang Dai}
\author[4]{Yujun Shi}
\author[5]{Xuan Ju}
\author[6]{Yushi Hu}
\author[2]{Artsiom Sanakoyeu}
\author[2]{Felix Juefei-Xu}
\author[2]{Ji Hou}
\author[2]{Junjiao Tian}
\author[2]{Tao Xu}
\author[2]{Tingbo Hou}
\author[2]{Yen-Cheng Liu}
\author[2]{Zecheng He}
\author[2]{Zijian He}
\author[3]{Matt Feiszli}
\author[2]{Peizhao Zhang}
\author[2]{Peter Vajda}
\author[2]{Sam Tsai}
\author[1]{Yun Fu}

\affiliation[1]{Northeastern University}
\affiliation[2]{Meta GenAI}
\affiliation[3]{Meta FAIR}
\affiliation[4]{National University of Singapore}
\affiliation[5]{The Chinese University of Hong Kong}
\affiliation[6]{University of Washington}

\contribution[*]{Work done at Meta}

\abstract{
Autoregressive (AR) models, long dominant in language generation, are increasingly applied to image synthesis but are often considered less competitive than diffusion-based models.
A primary limitation is the substantial number of image tokens required for AR models, which constrains both training and inference efficiency, as well as image resolution. 
To address this, we present Token-Shuffle, a novel yet simple method that reduces the number of image tokens in Transformers. Our key insight is the dimensional redundancy of visual vocabularies in Multimodal Large Language Models (MLLMs), where low-dimensional visual codes from the visual encoder are directly mapped to high-dimensional language vocabularies.
Leveraging this,  we consider two key operations: \textit{token-shuffle}, which merges spatially local tokens along channel dimension to decrease the input token number, and \textit{token-unshuffle}, which untangles the inferred tokens after Transformer blocks to restore the spatial arrangement for output. 
Jointly training with textual prompts, our strategy requires no additional pretrained text-encoder and enables MLLMs to support extremely high-resolution image synthesis in a unified next-token prediction framework while maintaining efficient training and inference. For the first time, we push the boundary of AR text-to-image generation to a resolution of $2048 \times 2048$ with impressive generation performance. In GenAI-benchmark, our 2.7B model achieves 0.77 overall score on hard prompts, outperforming AR models LlamaGen by 0.18 and diffusion models LDM by 0.15. Exhaustive large-scale human evaluations also demonstrate our superior image generation capabilities in terms of text-alignment, visual flaw, and visual appearance. We hope that Token-Shuffle can serve as a foundational design for efficient high-resolution image generation within MLLMs.
}
\correspondence{Xu Ma at \email{ma.xu1@northeastern.edu}}
\metadata[Project Page]{\url{https://ma-xu.github.io/token-shuffle/}}
\begin{document}

\maketitle
\begin{figure}
    \centering
    \includegraphics[width=0.96\textwidth]{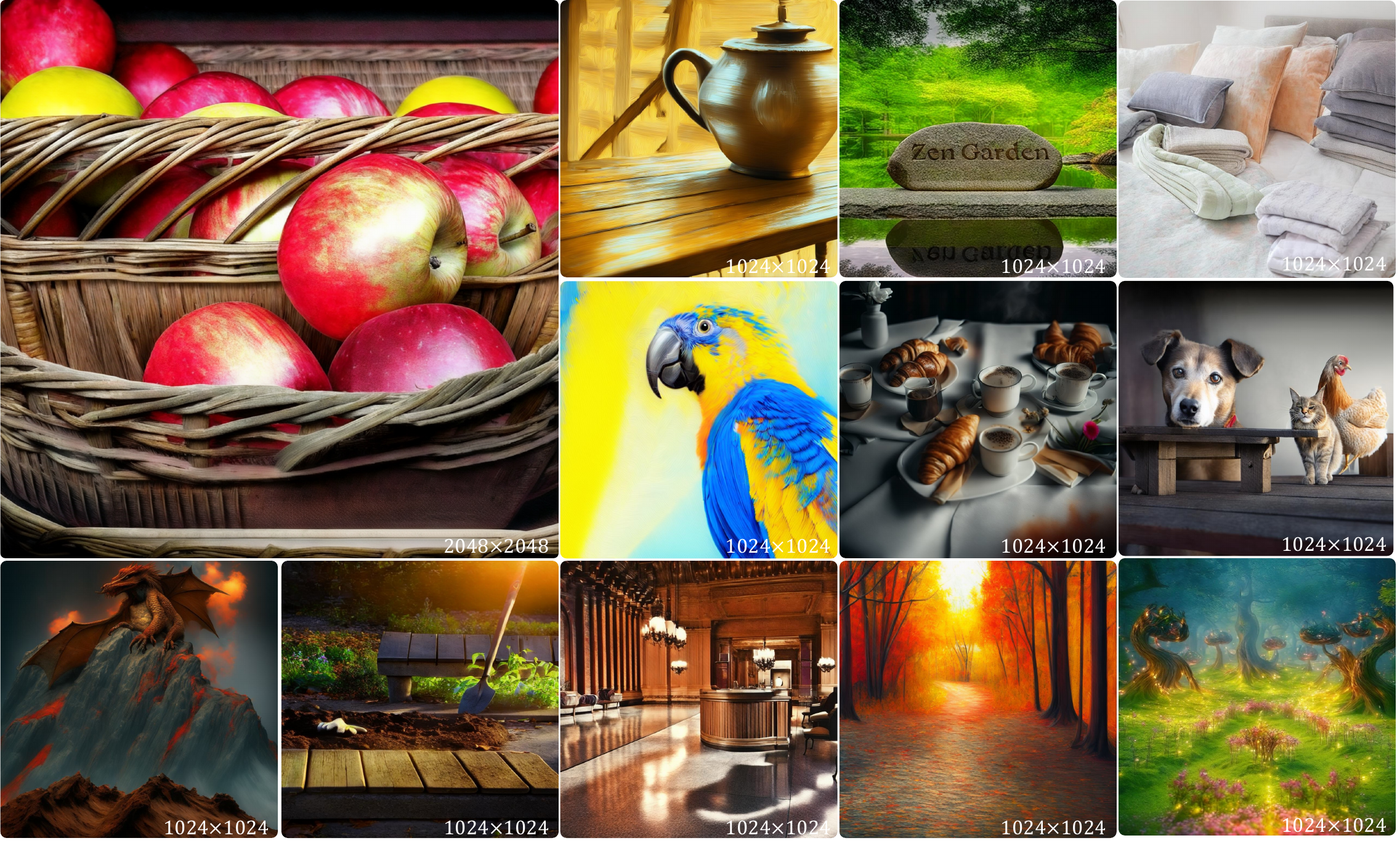}
    \vspace{-2mm}
    \captionof{figure}{
    High-resolution images generated by our 2.7B AR model with Token-Shuffle (shuffle window size = 2).
    }
    \label{fig:teaser}
    \vspace{-2mm}
\end{figure}
\section{Introduction}
Within the framework of autoregressive Transformers, large language models (LLMs)~\cite{touvron2023llama2, touvron2023llama, radford2018improving} have recently achieved remarkable success in natural language processing by predicting the next token in a sequence. Building on these advances, recent efforts have aimed to expand LLMs with image generation capabilities~\cite{pan2025transfer, sun2024emu, wang2024emu3,team2024chameleon}, leading to the development of multimodal large language models (MLLMs). 

Two primary strategies are explored for image generation in MLLMs: \textit{continuous} visual tokens~\cite{pan2025transfer,sun2024emu,li2024autoregressive} and \textit{discrete} visual tokens~\cite{sun2024emu, wang2024emu3, team2024chameleon, liu2024lumina}, each with unique pros and cons. Recent studies~\cite{fan2024fluid} highlight that continuous tokens deliver superior image quality and require fewer tokens, offering notable computational efficiency. In contrast, discrete tokens generally produce lower visual quality and require a quadratic increase in token count with respect to image resolution. However, discrete tokens are more compatible with LLMs, as they only require an expanded vocabulary size to accommodate visual vocabularies. Continuous tokens, on the other hand, necessitate extensive modifications to the LLM pipeline, including additional loss functions (\textit{e.g.}, regression~\cite{sun2024emu} or diffusion loss~\cite{li2024autoregressive}), adjustments to causal masking~\cite{li2024autoregressive, fan2024fluid}, and significant engineering efforts (\textit{e.g.}, model and loss parallelism). In addition, no strong evidence show that the continuous pipeline has less impact on text generation in MLLMs. Consequently, large-scale, real-world MLLM applications like EMU3~\cite{wang2024emu3} and Chameleon~\cite{team2024chameleon} predominantly adopt discrete visual tokens in practice.

Without altering the standard causal Transformers, discrete visual token MLLMs have explored applying the "next-token prediction" paradigm to image generation. Examples include LlamaGen~\cite{sun2024autoregressive}, Chameleon~\cite{team2024chameleon}, and EMU3~\cite{wang2024emu3}, which utilize vector quantization image tokenizers~\cite{van2017neural, esser2021taming} to convert images into discrete tokens, allowing autoregressive Transformers to generate images in a process similar to language generation. Although these MLLMs demonstrate impressive image generation capabilities, they face substantial limitations in terms of achievable resolution and the associated number of visual tokens. Unlike language, which typically requires a few dozen to a few hundred tokens, images demand far more—often thousands of tokens (\textit{e.g.}, 4K visual tokens to generate a $1024\times 1024$ resolution image).
Due to the quadratic computational complexity of Transformers, this huge token number requirement makes both training and inference prohibitively costly. As a result, most MLLMs are limited to generating low- or medium-resolution images~\cite{tian2024visual, sun2024autoregressive, wang2024emu3, liu2024lumina}, which restricts their ability to fully leverage the benefits of high-resolution images, such as enhanced detail preservation and fidelity. In contrast, high-resolution image generation has advanced significantly within the domain of diffusion models~\cite{ren2024ultrapixel, chen2024pixart, he2023scalecrafter, haji2023elasticdiffusion,qiu2024freescale}. While tentative efforts have been made towards efficient LLMs that support long-context generation (which also benefits high-resolution image generation), these typically involve architectural modifications~\cite{ding2023longnet, gu2023mamba, peng2023rwkv, katharopoulos2020transformers}, overlook off-the-shelf LLMs, or are optimized for language generation without leveraging the unique properties of images~\cite{gloeckle2024better}. Consequently, developing effective methods to scale image generation resolution with discrete visual tokens in MLLMs remains a key area of research.

To deal with this issue, we first look into the details of integrating visual tokens into the LLM vocabulary. As outlined above, the common practice is to concatenate the visual tokenizer codebook with the original LLM vocabulary to form a new multimodal vocabulary. While straightforward, this approach overlooks the intrinsic differences in dimensionality. For instance, in typical VQGAN implementations, the codebook vector dimension is relatively low, \textit{e.g.}, $256$~\cite{esser2021taming}. This low dimensionality is proven to be sufficient to distinguish vectors and has been shown to enhance both codebook usage and reconstruction quality~\cite{sun2024autoregressive,yu2021vector,yu2023language}. However, directly appending the visual tokenizer codebook to the LLM vocabulary results in a dramatic increase in vector dimension, reaching $3072$ or $4096$, or even higher. This drastic increase inevitably introduces ineffective dimensional redundancy for the added visual vocabulary, as we empirically demonstrated in Fig.~\ref{fig:motivation}.

\begin{wrapfigure}{r}{0.6\textwidth}
    \begin{center}
    \includegraphics[width=0.95\linewidth]{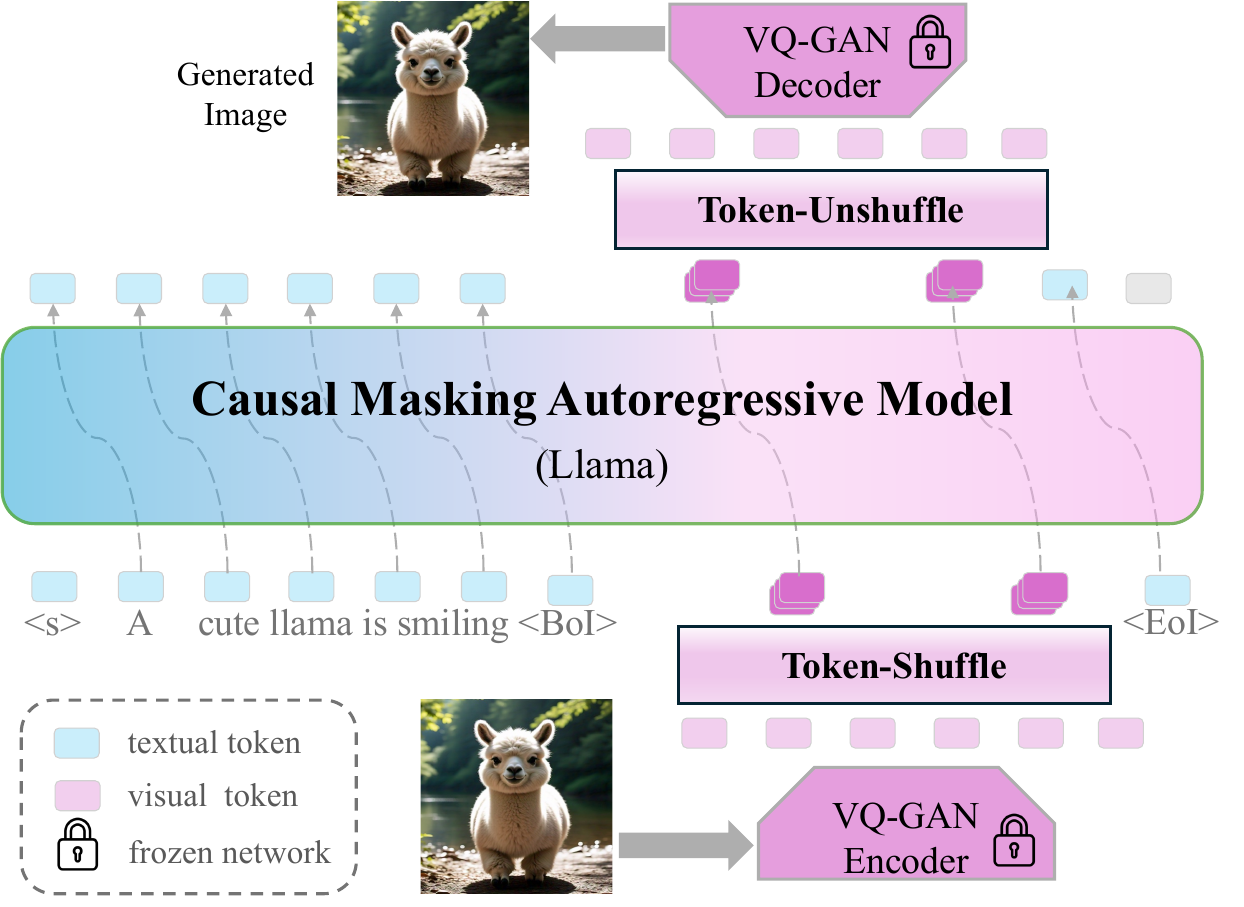}
    \end{center}
    \caption{
    \textbf{Token-Shuffle Pipeline:} a plug-and-play operation pair for reducing visual token number in MLLMs, comprising a token-shuffle operation to merge spatially local visual tokens for Transformer input and a token-unshuffle operation to disentangle inferred visual tokens. 
    }
    \label{fig:pipeline}
\end{wrapfigure}
Inspired by this, we introduce Token-Shuffle, a pair of plug-and-play operations designed for MLLMs that significantly reduces the number of visual tokens for computation, enhancing both efficiency and high-resolution image synthesis. Our method draws inspiration from the widely-used pixel-shuffle~\cite{shi2016real} technique in super-resolution, fusing visual tokens along the channel by leveraging the visual vocabulary dimensional redundancy. Rather than learning and generating each visual token individually, we process and generate a set of tokens within local windows sequentially, as illustrated in Fig.~\ref{fig:pipeline}.
This approach results in a \textit{drastic reduction} in the number of visual tokens for computation (\textit{e.g.}, saving $ \sim75\%$ of tokens when shuffle window size is set to 2) while maintaining high-quality generation.
Concurrent with our work, SynerGen-VL~\cite{li2024synergen} also explores Token Folding to reduce the number of visual tokens for both visual understanding and generation.
Fundamentally distinct from aggressive compression ratios on which conventional visual encoders rely, Token-Shuffle takes a novel approach by preserving fine-grained information and dynamically fusing tokens through the exploitation of visual dimensional redundancy. This strategy leads to enhanced image fidelity and details in generation. Comprehensive analysis is provided in Sec.~\ref{sec:compress_ratio}.

For the first time, Token-Shuffle pushes the boundaries of autoregressive image generation to a resolution of 2048$\times$2048 and makes it possible to go beyond, while still enjoying efficient training and inference. In addition to facilitating high-resolution image generation, Token-Shuffle preserves impressive generation quality. Using the 2.7B Llama model, we achieve an overall score of 0.62 on the GenEval benchmark~\cite{ghosh2024geneval} and a VQAScore of 0.77 on the GenAI-bench~\cite{li2024genai}, clearly outperforming related autoregressive models and even surpassing strong diffusion models, thereby setting a new state-of-the-art result. Besides, large-scale human evaluations also demonstrate the effectiveness of our proposed methods.
The effectiveness and efficiency of Token-Shuffle demonstrate the substantial potential of our method, empowering MLLMs with the capability for high-resolution, high-fidelity image generation and paving the way for surpassing diffusion-based approaches.

\section{Related Work}

\textbf{Text-to-Image Generation} aims to synthesize images based on class or textual prompts. Recently, diffusion-based models~\cite{ho2020denoising,song2020denoising,rombach2022high,peebles2023scalable,dai2023emu} have delivered impressive results.
Denoising diffusion probabilistic models (DDPM)~\cite{ho2020denoising_ddpm} laid the foundation of diffusion models, while denoising diffusion implicit models (DDIM)~\cite{ho2020denoising_ddim} introduced a more efficient and deterministic sampling process. 
Based on these, Latent diffusion models (LDM)~\cite{rombach2022high} innovatively shifted diffusion from pixel space to the latent space of powerful pretrained autoencoders, and introduced textual guidance. Other techniques, such as classifier-free guidance~\cite{ho2022classifier}, Flow Matching~\cite{lipman2022flow,polyak2024movie}, and v-prediction~\cite{salimans2022progressive}, have also contributed to improved image generation quality. 
Inspired by the success of Transformers in various tasks, recent approaches have explored Transformer designs for improved scalability, as demonstrated by models like DiT~\cite{peebles2023scalable} and U-ViT~\cite{bao2023all}. Moreover, work such as Imagen~\cite{saharia2022photorealistic} has demonstrated the effectiveness of leveraging large language models (LLMs) for image synthesis.
In our work, we take a different approach by directly exploring image synthesis using LLMs in an autoregressive manner, rather than diffusion-based methods.

\vspace{2mm}
\noindent\textbf{Autoregressive Models for Image Synthesis} have garnered significant attention recently. Unlike the dominant diffusion models, AR models offer the potential for a unified and general multimodal system.
One of the recent works is LlamaGen~\cite{sun2024autoregressive}, which employs a pure LLaMA~\cite{touvron2023llama} architecture to generate images via \textit{next-token prediction}. In contrast, the concurrent work VAR~\cite{tian2024visual} abandons next-token prediction in favor of next-scale prediction. 
Building on VAR, STAR~\cite{ma2024star} integrates an additional text encoder and introduces textual guidance through cross-attention. Meanwhile, Open-MAGVIT2~\cite{luo2024open} highlights the benefits of a visual tokenizer with an extensive vocabulary.
In a different approach, MAR~\cite{li2024autoregressive} eliminates the need for discrete visual tokens and instead uses a lightweight diffusion block to decode continuous latent features.
However, the above approaches either focus on class-conditioned synthesis within predefined categories or rely on additional pretrained and frozen text encoders for text-conditioned synthesis. A unified autoregressive MLLM for text-conditioned image generation remains underexplored, and this is the focus of our work.

\vspace{2mm}
\noindent\textbf{Multimodal Large Language Models} are designed to understand and generate across various modalities~\cite{yu2023scaling}. Given the recent wave of successes with LLMs~\cite{mann2020language,dubey2024llama}, it is natural to extend LLMs into the multimodal domain. In such models, different modalities are encoded via specific tokenizers, fused, and jointly learned with other modalities.  
Conceptually, recent advances in multimodal models generally fall into two approaches: one uses continuous tokens for non-text modalities, and the other is based on discrete token representations for all modalities. For approaches with continuous tokens, they incorporate continuous features like VAE or CLIP~\cite{radford2021learning} features of visual data into LLMs for improved multimodal understanding and generation. These methods often result in better generation quality compared to discrete token-based models. As a result, numerous models have emerged, including EMU~\cite{sun2024emu}, EMU2~\cite{sun2024generative}, SEED-X~\cite{ge2024seed}, FLUID~\cite{fan2024fluid}, and MetaQueries~\cite{pan2025transfer}, \textit{etc}.
On the other hand, one of the leading models in the discrete token representation category is CM3Leon~\cite{yu2023scaling}, which builds on the CM3~\cite{aghajanyan2022cm3} multimodal architecture. In addition to text and image generation and infilling, CM3Leon demonstrates the strong benefits of scaling and fine-tuning on diverse instruction-based datasets. Similar models, such as Chameleon~\cite{team2024chameleon}, EMU3~\cite{wang2024emu3} and Lumina-mGPT~\cite{liu2024lumina}, have also shown promising results.
In our work, we consider discrete tokens for MLLM image generation and target efficient high-resolution image generation. 

\section{Token-Shuffle}
We present Token-Shuffle, a straightforward yet powerful method for reducing the number of visual tokens in MLLMs, significantly lowering computational costs and enabling efficient high-resolution image synthesis.

\subsection{Preliminary} 
\noindent\textbf{Large Language Model Architecture} Our approach utilizes a decoder-only autoregressive Transformer model, specifically LLaMA~\cite{dubey2024llama}, as the foundational language model. 
LLMs like LLaMA model the conditional probability of the $t$-th token $\mathbb{P}\left(x_t | x_1, x_2, \cdots, x_{t-1}\right)$ through an autoregressive \textit{next-token prediction} process, and only require the standard cross-entropy loss for training. 

\noindent\textbf{Image Synthesis in LLMs} To enable LLMs to perform image synthesis, we incorporate discrete visual tokens into the model's vocabulary. We utilize the pretrained VQGAN model from LlamaGen~\cite{sun2024autoregressive}, which down-samples the input resolution by a factor of 16. The VQGAN codebook contains 16,384 tokens, which are concatenated with LLaMA's original vocabulary. Additionally, special tokens \texttt{<|start\_of\_image|>} and \texttt{<|end\_of\_image|>} are introduced to encapsulate the sequence of discrete visual tokens. During training, all tokens—including visual and textual—are used to compute the loss.

\subsection{Limitations for Image synthesis} While various models have demonstrated the ability of image synthesis in MLLMs by inferring discrete visual tokens~\cite{sun2024autoregressive,yu2023scaling,wang2024emu3}, an inevitable issue is the prohibitive number of visual tokens for high-resolution images. As aforementioned, to generate a high-resolution image with a resolution of $1024\times 1024$, it requires a total of $\mathbf{4K}$ visual tokens if a down-sample 16 tokenizer is employed. Compared to language corpora, such a number of visual tokens makes the training extremely slow and the inference prohibitively inefficient. This will also largely restrict the generated image quality and aesthetic~\cite{sun2024autoregressive,rombach2022high}. 
Moreover, if we increase the resolution to $2048\times 2048$, it will corresponding increase to $\mathbf{16K}$, which is impractical for both effective training and efficient inference in the context of \textit{next-token-prediction}.

In principle, increasing the number of visual tokens can yield more detailed, aesthetically pleasing images with higher resolution. However, this also introduces a prohibitive computational and inference burden. Previous approaches have always faced the trade-off: either enduring significantly increased training and inference costs, or sacrificing image resolution and quality. Addressing this dilemma is of particular interest in the field, as people seek methods that balance efficiency with high-fidelity image generation.

\subsection{Visual Dimensional Redundancy}

\begin{figure}[!h]
    \centering
    \includegraphics[width=0.3\linewidth]{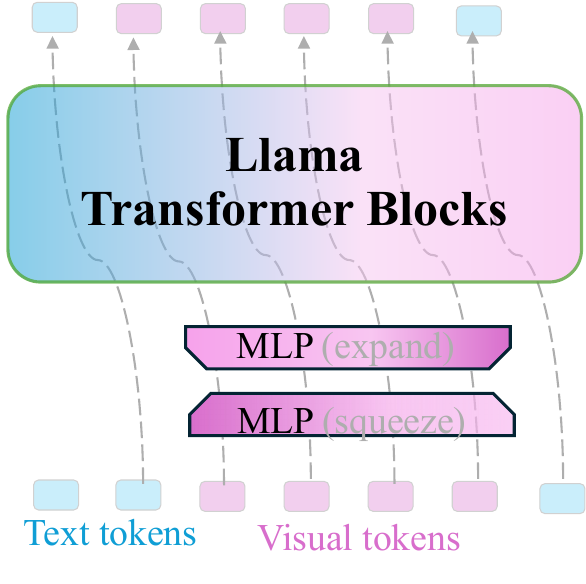}
    \hspace{10mm}
    \includegraphics[width=0.4\linewidth]{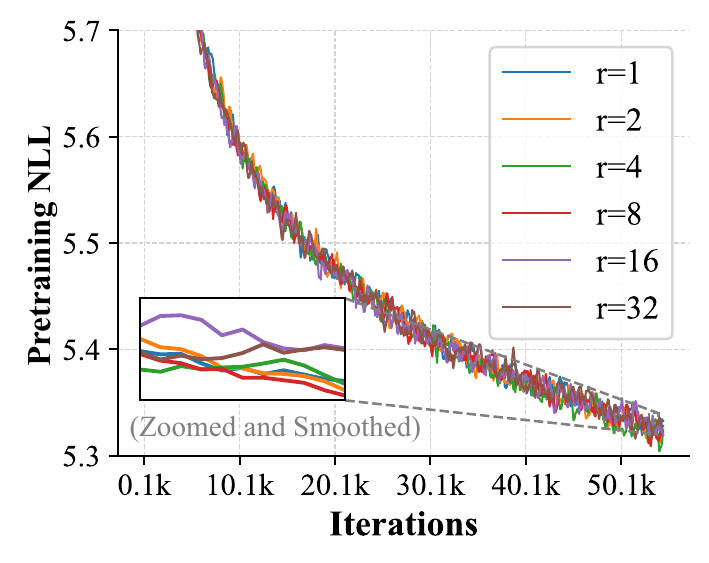}
    \vspace{-2mm}
    \caption{
    \textbf{Illustration of visual vocabulary dimensional redundancy.} \textbf{Left:} Two MLPs reduce visual token rank by a factor of $r$. \textbf{Right:} Pre-training loss (log-scaled perplexity) for different $r$ values, showing substantial dimension reduction with minimal performance impact.
    }
    \label{fig:motivation}
\end{figure}

As illustrated above, a common strategy to endow large language models (LLMs) with image generation capabilities involves appending visual codebook tokens to the language vocabulary. While conceptually straightforward, this approach leads to a substantial increase in the embedding dimensionality for visual tokens.
We contend that such a common approach of directly incorporating discrete visual tokens into the vocabulary of LLMs introduces inherent dimensional redundancy. To investigate this, we conduct a simple study using a 2.7B Llama-based MLLM with a dimension of 3072. For visual vocabularies, we introduce two linear layers to linearly reduce and expand the embedding dimension. This configuration ensures that the rank of the visual vocabulary is constrained to $< \frac{3072}{r}$, where $r$ is the compression factor. We train models with varying values of $r$ on a licensed dataset for 55K iterations for demonstration.
Fig.~\ref{fig:motivation} shows that there is considerable redundancy in visual vocabularies, and we can compress the dimension by up to a factor of 8 without significantly impacting generation quality. A slight increase in loss is observed when larger compression factors are used.

\subsection{Token-Shuffle Operations}

Motivated by our analysis of dimensional redundancy in visual vocabularies, we introduce \textit{\textbf{Token-Shuffle}}—plug-and-play operations that reduce visual token counts in Transformer to improve computational efficiency and enable high-resolution image generation.

\begin{figure}
    \centering
    \includegraphics[width=0.8\linewidth]{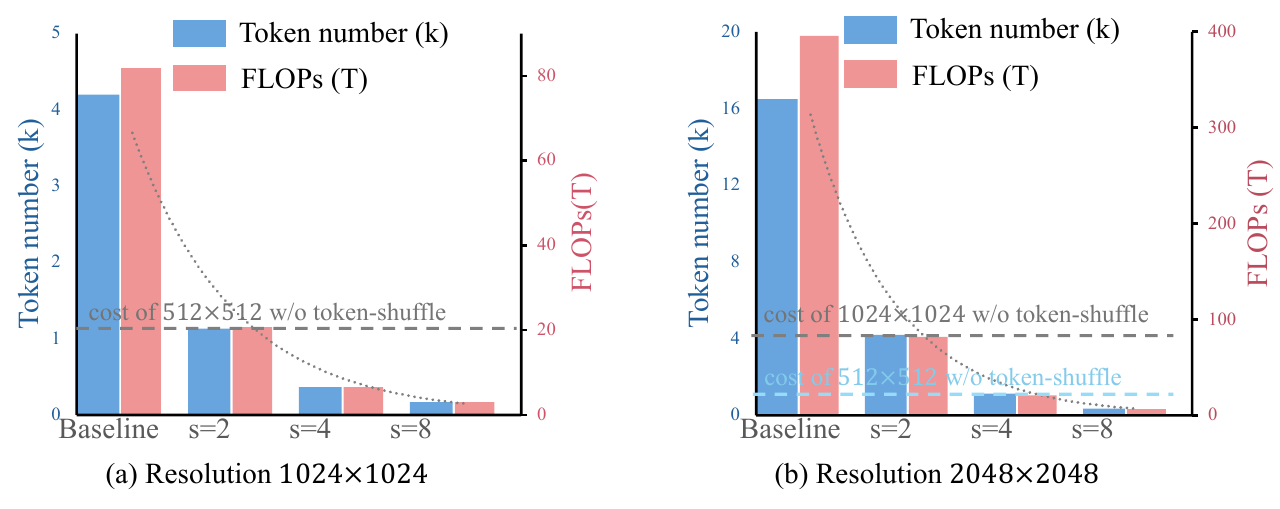}
    \caption{\textbf{Token-Shuffle can enhance efficiency quadratically}. For instance, with a shuffle window size $s=2$, we achieve approximately a $4\times$ reduction in both training FLOPs and token number. Considering the use of KV-cache during inference, inference time scales roughly linearly with the token number.}
    \label{fig:efficiency}
    \vspace{-3mm}
\end{figure}

\textit{
Rather than reducing dimensional redundancy of visual vocabulary, we leverage this redundancy to reduce the number of visual tokens for greater efficiency}. Specifically, we shuffle spatially local visual tokens into a single token, then feed these fused visual tokens along with textual tokens into Transformer. A shared MLP layer is employed to compress visual token dimension, ensuring  the fused token has same dimension as original. Assuming a local shuffle window size of $s$, our method reduces the token number by a factor of $s^2$, significantly alleviating the computational burden on the Transformer architecture.
To recover the original visual tokens, we further introduce a token-unshuffle operation that disentangles the fused tokens back into their local visual tokens, with additional MLP layer to restore the original dimensionality. We also introduce residual MLP blocks in both operations. The entire Token-Shuffle pipeline is illustrated in Fig.~\ref{fig:pipeline} for clarity. \textit{In essence, we do not reduce the number of tokens during inference or training but instead reduce the token count during Transformer computation.} 
Fig.~\ref{fig:efficiency} illustrates the efficiency of our proposed method.
Moreover, rather than strictly adhering to the \textit{next-token-prediction} paradigm, our approach predicts the \textit{next fused token}, allowing us to output a set of local visual tokens in a single step, which significantly improves the efficiency and makes the high-resolution image generation feasible for AR models. See supplementary for analysis on causal attention. 

\subsection{Token-Shuffle Implementation Details}
For Transformer input, we first compress the visual vocabulary dimension by a factor of ${s^2}$ via an MLP layer that maps the dimension from $d$ to $\frac{d}{s^2}$, where $d$ represents the Transformer dimension. Next, local $s \times s$ visual tokens are shuffled into a single token, reducing the total number of tokens per image from $n$ to $\frac{n}{s^2}$ while preserving the overall dimensionality. To enhance visual feature fusion, we add $n$ MLP blocks.
For Transformer output, we employ Token-Unshuffle. Here, MLP blocks map features into a new space, and an unshuffle operation expands each output visual token back to $s \times s$ tokens. Another MLP layer then restores the dimension from $\frac{d}{s^2}$ to $d$, with additional MLP blocks used to refine feature extraction. Consistently, both Token-Shuffle and Token-Unshuffle utilize $n$ MLP layers for simplicity, where each MLP block consists of two linear projections with GELU activation. Further design choices for Token-Shuffle are explored in Sec~\ref{sec:token_shuffle_design_choice}.

\begin{figure*}
    \centering
    \includegraphics[width=1\linewidth]{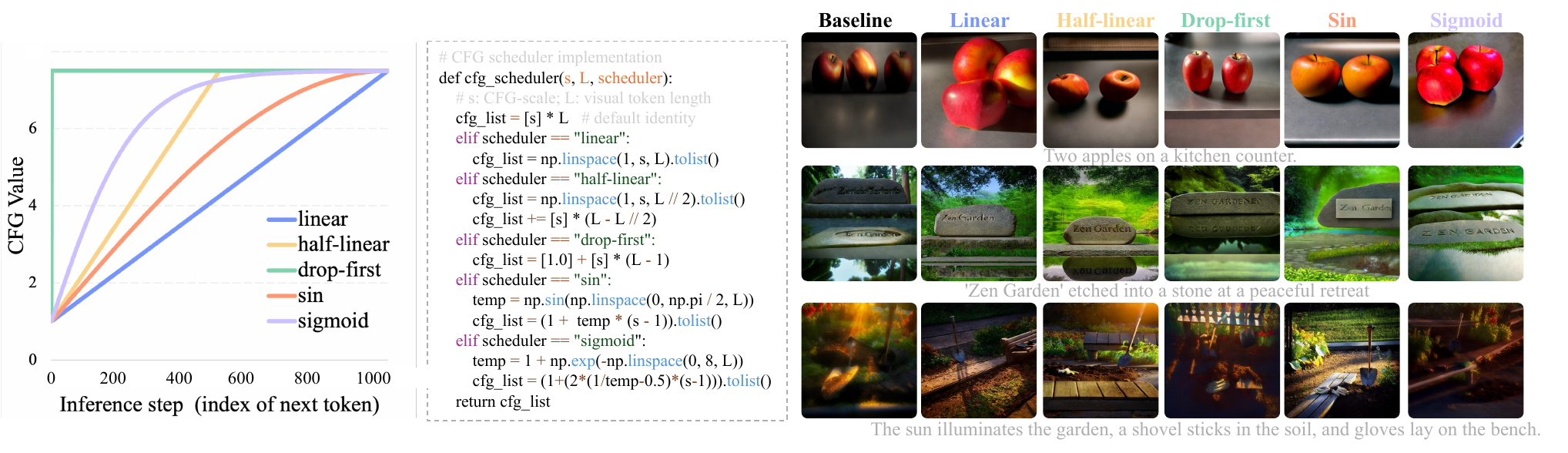}
    \vspace{-6mm}
    \caption{\textbf{Comparison of different CFG schedulers} with a monotonic increase in CFG scale from 1 to 7.5. \textbf{Right:} CFG-scheduler improves both visual aesthetics and text alignment, compared to the baseline of a consistent CFG value of 7.5 across all visual tokens.}
    \label{fig:cfg_scheduler}
    \vspace{-5mm}
\end{figure*}

\subsection{CFG Scheduler for AR Image Generation}
Following common practice~\cite{sun2024autoregressive,wang2024emu3}, we incorporate classifier-free guidance (CFG)~\cite{ho2022classifier} during both training and inference, a technique widely used in the Diffusion community. During training, we randomly drop 10\% of prompts, making the unconditional input format {\small \texttt{<|begin\_of\_sentence|><|begin\_of\_image|> ... <|end\_of\_image|><|end\_of\_sentence|>}}. In inference, we adjust the logits of each visual token as $l = l_{uncond} + \alpha(l_{cond} - l_{uncond})$ sequentially, where $\alpha$ is a hyperparameter that controls the text-image alignment.

However, AR-based models differ fundamentally from diffusion-based models, and we argue that the vanilla CFG implementation may not be optimal for AR models. For unconditional input, generated image tokens are consistently conditioned on two system tokens, {\small \texttt{<|begin\_of\_sentence|>}} and {\small \texttt{<|begin\_of\_image|>}}. 
That is, the first unconditional logits always remain consistent, and applying the first fixed logits to conditional input logits may introduce unpredictable artifacts. These small errors accumulate auto-regressively from the first to the last token, potentially resulting in degraded image quality. 
Inspired by recent work~\cite{wang2024analysis}, we further introduce a new inference CFG-scheduler to improve image generation quality. Our motivation is to minimize, or even eliminate, the influence of unconditional logits on early visual tokens to prevent artifacts. The cumulative impact of CFG from the first to last token would be enough to enhance both image quality and adherence to conditions. We explored several CFG-scheduler strategies, with results presented in Fig.~\ref{fig:cfg_scheduler} (zoom in for better visualization). Suggested by visual quality and human evaluation, we consider the half-linear scheduler for better generation by default.

\section{Experiments}
\subsection{Training Details}
We conduct all experiments using the 2.7B Llama model, which has a dimension of 3072 and consists of 20 autoregressive Transformer blocks. The models are trained on licensed dataset following Emu~\cite{dai2023emu}. For training high-resolution images at $2048\times 2048$, we exclude images with a resolution smaller than $1024\times 1024$. Our model is initialized with the text pretrained 2.7B Llama checkpoint and begins training with a learning rate of $2e^{-4}$. All image captions are rewritten by Llama3~\cite{dubey2024llama} to generate long prompts with details, which is demonstrated to be helpful for better generation.

We pre-train the models in three stages, from low-resolution to high-resolution image generation. 
First, we train the models using an image resolution of $512\times 512$ without employing the Token-Shuffle operation, as the number of visual tokens is not substantial at this stage. In this stage, we train on approximately 50 Billion tokens, using a sequence length of 4K, a global batch size of 512, and a total of 211K steps.
Next, we scale the image resolution up to $1024\times 1024$ and introduce the Token-Shuffle operation to reduce the number of visual tokens for improved computational efficiency. In this stage, we scale up to 2 TB training tokens. 
Finally, we further scale up to $2048\times 2048$ using the previously trained checkpoint on around 300 Billion tokens with an initial learning rate of $4e^{-5}$. Unlike training on lower resolutions, \textit{we observe that handling higher resolutions (e.g., $2048\times 2048$) always results in unstable training, with the loss and gradient value increasing unexpectedly.} To address this, we incorporate z-loss~\cite{team2024chameleon}, which stabilizes training for very-high-resolution image generation. Detailed exploration and implementation specifics are provided in supplementary Sec.~\ref{sec:implementation_details}.
We fine-tune all models at different stages with a learning rate of $4e^{-6}$ on 1,500 selected high-aesthetic quality images for presentation. By default, we present visualizations and evaluations based on the fine-tuned results at a resolution of $1024\times 1024$ and token-shuffle window size of 2, unless otherwise specified.

\subsection{Quantitative Evaluation}

\begin{table*}
\centering
\resizebox{0.99\linewidth}{!}{
\begin{tabular}{lcccccccccccccc}
\toprule
\multirow{3}{*}{\textbf{Model}} & \multirow{3}{*}{\textbf{Type}}&\multicolumn{6}{c}{\textbf{"Basic" prompts}}&\color{white}{.} & \multicolumn{6}{c}{\textbf{"Hard" prompts}}  \\
\cmidrule{3-8} \cmidrule{10-15}
 & &\multirow{2}{*}{\textbf{Attribute}}& \multirow{2}{*}{\textbf{Scene}} & \multicolumn{3}{c}{\textbf{Relation}} & \multirow{2}{*}{\textbf{Overall}} &  &\multirow{2}{*}{\textbf{Count}} & \multirow{2}{*}{\textbf{Differ}} & \multirow{2}{*}{\textbf{Compare}}& \multicolumn{2}{c}{\textbf{Logical}} & \multirow{2}{*}{\textbf{Overall}}\\
\cmidrule{5-7} \cmidrule{13-14}
 & & & & Spatial & Action & Part & &  & & & & Negate & Universal &\\
\midrule
SDXL-v2.1       & Diff.& 0.80 & 0.79 & 0.76 & 0.77 & 0.80 & 0.78 & & 0.68 & 0.70 & 0.68 & 0.54 & 0.64 & 0.62\\
SD-XL Turbo     & Diff.& 0.85 & 0.85 & 0.80 & 0.82 & 0.89 & 0.84 & & 0.72 & 0.74 & 0.70 & 0.52 & 0.65 & 0.65\\
DeepFloyd-IF~\cite{saharia2022photorealistic}    &Diff. & 0.83 & 0.85 & 0.81 & 0.82 & 0.89 & 0.84 & & 0.74 & 0.74 & 0.71 & 0.53 & 0.68 & 0.66\\
Midjourney v6   & Diff.& 0.88 & 0.87 & 0.87 & 0.87 & 0.91 & 0.87 & & 0.78 & 0.78 & 0.79 & 0.50 & 0.76 & 0.69\\
DALL-E 3~\cite{betker2023improving}        &Diff. & 0.91 & 0.90 & 0.92 & 0.89 & 0.91 & \textbf{0.90} & & 0.82 & 0.78 & 0.82 & 0.48 & 0.80 & 0.70 \\
\midrule
LlamaGen~\cite{sun2024autoregressive} &AR & 0.75 & 0.75 & 0.74&0.76 &0.75 &0.74 &  &0.63 & 0.68 &0.69  &0.48 &0.63 & 0.59  \\
Lumina-mGPT-7B~\cite{liu2024lumina}&AR & 0.84 &0.85  &0.82&0.84 &0.93 &0.83 &  &0.75 &0.69  &0.73  &0.47 &0.69 &0.63 \\
EMU3~\cite{wang2024emu3} &AR & 0.78 &0.81  &0.77 &0.78 &0.87  &0.78  &  & 0.69 & 0.62  & 0.70  & 0.45 &0.69  & 0.60   \\
SEED-X~\cite{ge2024seed} &AR+Diff. &0.86  & 0.88 &0.85 &0.85 &0.90  &0.86 &  &0.79  & 0.77  &0.77   & 0.56 &0.73  & 0.70\\
Token-Shuffle &AR & 0.78 &0.77  &0.80 &0.76 &0.83 &0.78 & &0.76  &0.74 &0.74 &0.58 & 0.64&0.67  \\
Token-Shuffle\textcolor{blue}{${\dagger}$} &AR &0.88 & 0.88 &0.88 &0.87 &0.91 & 0.88 & & 0.81 &0.82 & 0.81 &0.68  &0.78 &\textbf{0.77}\\
\bottomrule
\end{tabular}
}
\caption{\textbf{VQAScore evaluation of image generation on GenAI-Bench.} "\textcolor{blue}{$\dagger$}" indicates that images are generated by Llama3-rewritten prompts to match the caption length in the training data, for training-inference consistency.}
\label{tab:vqascore}
\end{table*}

\begin{table}
    \centering
    \resizebox{0.99\linewidth}{!}{
    \begin{tabular}{lccccccccc}
        \toprule
        \textbf{Method} &\textbf{Type}& \textbf{\# Params} & \textbf{Single Obj.} & \textbf{Two Obj.} & \textbf{Counting} & \textbf{Colors} & \textbf{Position} & \textbf{Color Attri.} & \textbf{Overall $\uparrow$} \\
        \midrule
        LDM~\cite{rombach2022high}  &Diff. & 1.4B & 0.92 & 0.29 & 0.23 & 0.70 & 0.02 & 0.05 & 0.37 \\
        SDv1.5~\cite{rombach2022high}   &Diff. & 0.9B & 0.97 & 0.38 & 0.35 & 0.76 & 0.04 & 0.06 & 0.43 \\
        PixArt-alpha~\cite{chen2024pixart}&Diff. & 0.6B & 0.98 & 0.50 & 0.44 & 0.80 & 0.08 & 0.07 & 0.48 \\
        SDv2.1~\cite{rombach2022high}  &Diff. & 0.9B & 0.98 & 0.51 & 0.44 & 0.85 & 0.07 & 0.17 & 0.50 \\
       DALL-E 2~\cite{ramesh2022hierarchical}   &Diff.   & 6.5B & 0.94 & 0.66 & 0.49 & 0.77 & 0.10 & 0.19 & 0.52 \\
        SDXL~\cite{podell2023sdxl}   &Diff.  & 2.6B & 0.98 & 0.74 & 0.39 & 0.85 & 0.15 & 0.23 & 0.55 \\
        SD3~\cite{esser2024scaling} &Diff. & 2B   & 0.98 & 0.74 & 0.63 & 0.67 & 0.34 & 0.36 & 0.62 \\
        \midrule
        Show-o~\cite{xie2024show} &AR.+Diff. & 1.3B &0.95&0.52&0.49&0.82&0.11&0.28&0.53\\
        SEED-X~\cite{ge2024seed} &AR.+Diff. & 17B & 0.97 & 0.58 & 0.26  & 0.80  & 0.19  & 0.14  & 0.49 \\
        Transfusion~\cite{zhou2024transfusion} &AR.+Diff. & 7.3B & -& -& -& -& -& -&0.63 \\
        LlamaGen~\cite{sun2024autoregressive}  &AR.   & 0.8B & 0.71 & 0.34 & 0.21 & 0.58 & 0.07 & 0.04 & 0.32 \\
        Chameleon~\cite{team2024chameleon} &AR. & 7B & -& -& -& -& -& -&0.39 \\
        EMU3~\cite{wang2024emu3}&AR.& 8B & -& -& -& -& -& -&0.66 \\
        EMU3-DPO~\cite{wang2024emu3}&AR.& 8B & -& -& -& -& -& -&0.64 \\
        Emu3-Gen~\cite{wang2024emu3}&AR.& 8B &0.98 &0.71 &0.34 &0.81 &0.17 &0.21& 0.54\\
        Janus~\cite{wu2024janus}&AR. & 1.3B &0.97 &0.68 &0.30 &0.84 &0.46 &0.42 &0.61\\
        Token-Shuffle\textcolor{blue}{${\dagger}$}&AR. &2.7B & 0.96 & 0.81 & 0.37 & 0.78 & 0.40 & 0.39 & 0.62 \\
        \bottomrule
    \end{tabular}
    }
    \caption{\textbf{Evaluation on the GenEval benchmark.} Similar to ours results, EMU3 and EMU3-DPO also consider prompt rewriting, and results of EMU3-Gen are from Janus~\cite{wu2024janus}. These results indicate our Token-Shuffle can also present promising generation quality besides high-resolution.}
    \label{tab:geneval}
    \vspace{-3mm}
\end{table}

\begin{figure*}[]
    \centering
    \includegraphics[width=0.99\linewidth]{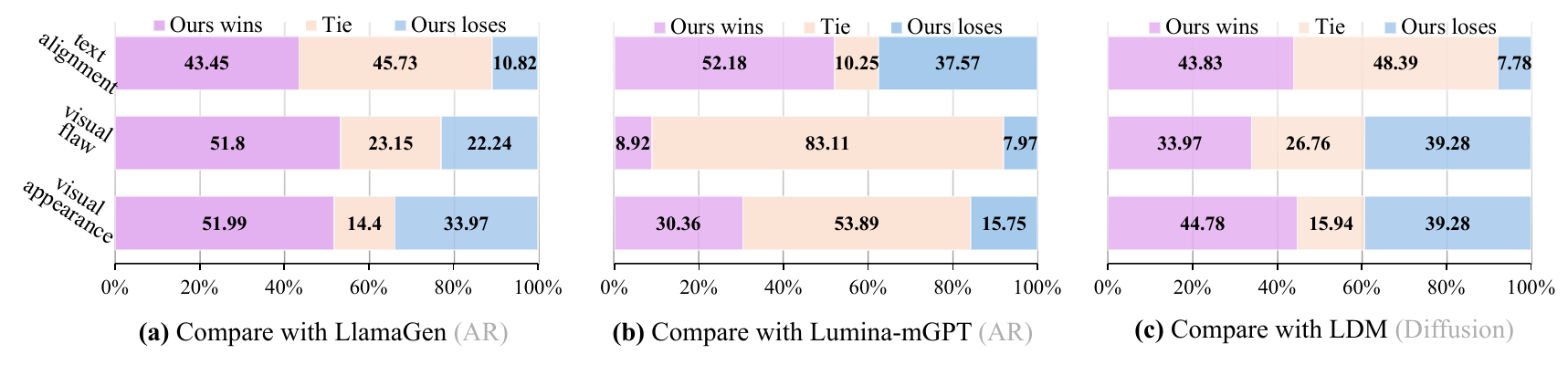}
\caption{\textbf{Human evaluation} comparing Token-Shuffle with LlamaGen~\cite{sun2024autoregressive} (AR-based model without text), Lumina-mGPT~\cite{liu2024lumina} (AR-based model with text) and LDM~\cite{rombach2022high} (diffusion-based model) on text alignment, visual flaws, and visual appearance.}
    \label{fig:human_eval}
    \vspace{-4mm}
\end{figure*}

While FID~\cite{heusel2017gans} or CLIPScore~\cite{hessel2021clipscore} are commonly used for image generation evaluation for class-conditioned synthesis, it is well-known that metrics are not reasonable for textual guided generation, as demonstrated in various related works~\cite{lin2024evaluating,ghosh2024geneval}. In our work, we consider two benchmarks: GenEval~\cite{ghosh2024geneval} and GenAI-Bench~\cite{li2024genai}. GenAI-Bench uses VQAScore~\cite{lin2024evaluating} as the auto-evaluation metric, which fine-tuned a visual-question-answering (VQA) model to produce an text-image alignment score. 
Since our training captions are long captions similar to LlamaGen~\cite{sun2024autoregressive}, we report results based on Llama3-rewritten prompts for caption length consistency. Additionally, we include results from the original prompts for reference. 

The results in Tab.~\ref{tab:vqascore} highlight the strong performance of our Token-Shuffle. Compared with other autoregressive models, our method outperforms LlamaGen by an overall score of 0.14 on "basic" prompts and 0.18 on "hard" prompts. Against strong diffusion-based baselines, our method surpasses DALL-E 3 by 0.7 in overall score on "hard" prompts.

Besides VQAScore results reported in Table~\ref{tab:vqascore}, we also conduct additional auto-evaluation, GenEval, and report the detailed evaluation results in Table~\ref{tab:geneval}. All inference configurations are the same and we consider the rewritten prompt by default.
Experimental results indicate that besides high-resolution, our Token-Shuffle, a pure AR-model, is able to present promising generation quality.

\subsection{Human Evaluation}

\begin{wrapfigure}{r}{0.6\textwidth}
    \centering
    \includegraphics[width=0.99\linewidth]{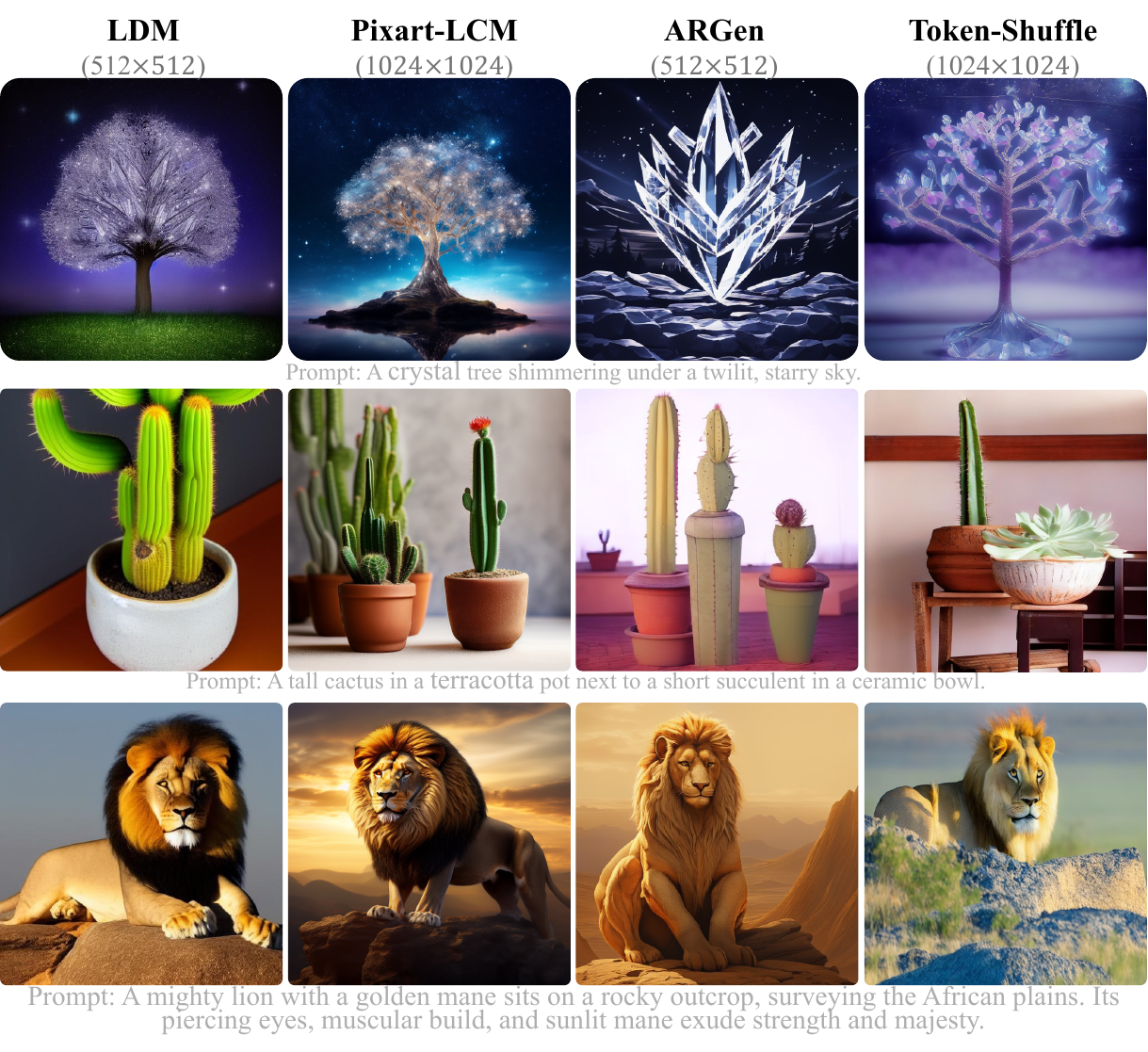}
    \vspace{-3mm}
    \caption{\textbf{Visual comparison} with other open-source diffusion-based and AR-based models (zoom in for details).}
    \label{fig:visuial_compare}
    \vspace{-4mm}
\end{wrapfigure}
We recognize that while automated evaluation metrics provide unbiased assessments, they may not always fully capture human preferences, as suggested by recent studies~\cite{dai2023emu, kirstain2023pick, podell2023sdxl}. To this end, we also conducted large-scale human evaluations on the GenAI-bench prompts set, comparing our model with LlamaGen~\cite{sun2024autoregressive}, Lumina-mGPT~\cite{liu2024lumina}, and with LDM~\cite{rombach2022high}, as representative methods for AR model, MLLM, and Diffusion, respectively.
For human evaluation, we focus on three key metrics: \textbf{text alignment}, assessing the accuracy with which images reflect textual prompts; \textbf{visual flaws}, checking for logical consistency to avoid issues such as incomplete bodies or extra limbs; and \textbf{visual appearance}, which evaluates the aesthetic quality of the images. 

Fig.~\ref{fig:human_eval} presents the results, where our model consistently outperforms AR-based model LlamaGen and Lumina-mGPT across all evaluation aspects. This suggests that Token-Shuffle effectively preserves aesthetic details and closely adheres to textual guidance with adequate training, even when token count is largely reduced for efficiency. In comparison with LDM, we demonstrate that AR-based MLLMs can achieve comparable or superior generation results (in terms of both visual appearance and text alignment) relative to Diffusion models. However, we observe that Token-Shuffle performs slightly worse than LDM in terms of visual flaws, consistent with observations in Fluid~\cite{fan2024fluid}, highlighting an interesting area for further exploration.

\subsection{Visual Examples}
We compare Token-Shuffle visually against other models, including two diffusion-based models, LDM and Pixart-LCM~\cite{chen2024pixart}, and one autoregressive model, LlamaGen~\cite{sun2024autoregressive}. The visual examples are presented in Fig.~\ref{fig:visuial_compare}. While all models exhibit favorable generation results, our Token-Shuffle appears to align more closely with the text, as demonstrated in rows 4 and 5. A possible reason for this is that we jointly train text and image within a unified MLLM-style model. Compared to AR model LlamaGen, Token-Shuffle achieves higher resolution at the same inference cost, offering improved visual quality and text alignment. When compared to diffusion-based models, our AR-based model Token-Shuffle demonstrates competitive generation performance, while also supporting high-resolution outputs.

\subsection{Ablation study}

\subsubsection{Design choice of Token-Shuffle} 
\label{sec:token_shuffle_design_choice}
\begin{wrapfigure}{r}{0.55\textwidth}
    \centering
    \vspace{-5mm}
    \includegraphics[width=1\linewidth]{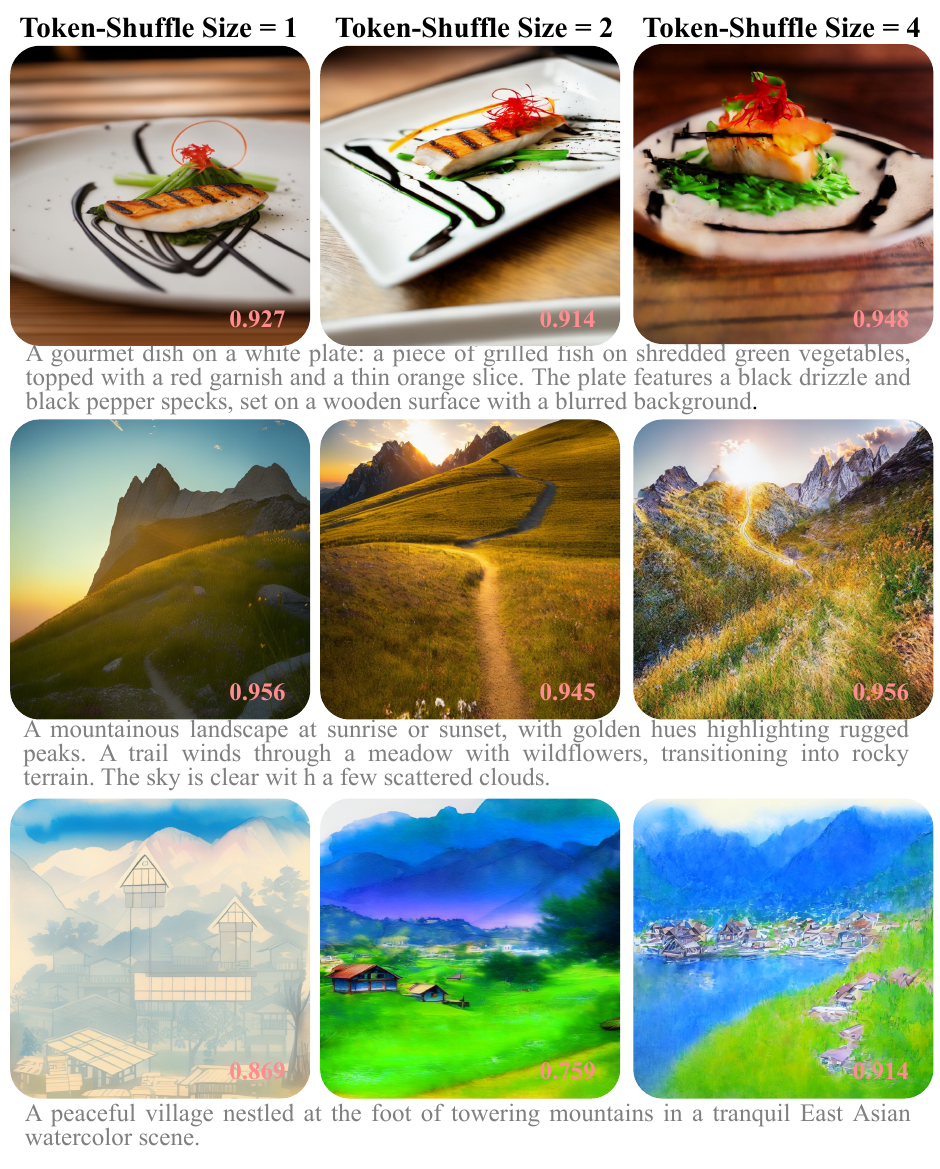}
    \vspace{-6mm}
    \caption{Visual comparison of different Token-Shuffle window sizes. We tested each prompt with fixed random seeds and reported the VQAScore~\cite{lin2024evaluating} in the bottom-right corner.
    }
    \label{fig:visualization_s1s2s4}
    \vspace{-6mm}
\end{wrapfigure}
We acknowledge that similar implementations of Token-Shuffle or alternative methodologies may also be effective. Here, we explore and evaluate several variations:
\begin{itemize}
    \item \textit{More MLP blocks.} In the default setting, we use $n=2$ MLP blocks. To assess whether increasing the number of MLP blocks enhances performance, we experiment with configurations of $n=4$ and $n=6$.
    \item \textit{Shuffle or Drop.} To determine the importance of each token within local windows, we compare the standard Token-Shuffle operation with a variation in which all tokens in a local window are dropped except the last one.
    \item \textit{Additional Positional Embedding.} 
    As MLP layers are position-aware, we do not include additional positional embeddings in the default setup, and RoPE is already used in the Transformer blocks. To evaluate the potential benefits of additional positional embeddings, we introduce learnable embeddings at the local (shared and within shuffle-window) and global ranges, respectively.

    \item \textit{Re-sampler and Simple version.} We further explore re-sampler~\cite{ge2024seed} to fuse and decouple tokens, replacing the Token-Shuffle design. In addition, we follow the common practice for high-resolution image understanding in Vision-Language Models, which directly concatenates local visual features and use MLP to match dimension. For outputs, we first use an MLP to expand the dimension and then decouple the tokens. We term this option as simple version. Notice that all operations in simple version are linear. 
\end{itemize}

\begin{figure*}[]
    \centering
    \begin{subfigure}[b]{0.24\linewidth}
        \includegraphics[width=0.99\linewidth]{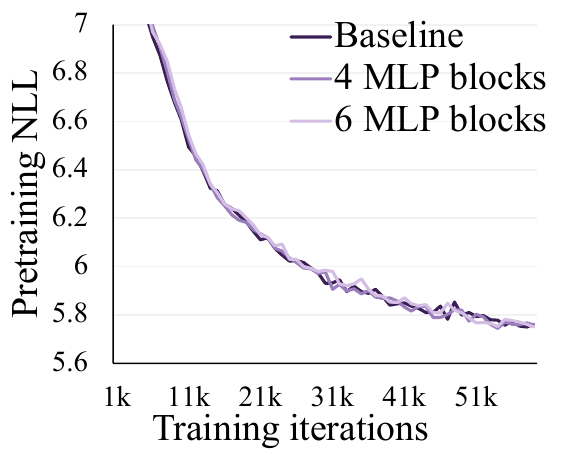}
        \caption{More MLP blocks}
        \label{fig:ablation_depth}
    \end{subfigure}
    % \quad
    \begin{subfigure}[b]{0.24\linewidth}
        \includegraphics[width=0.99\linewidth]{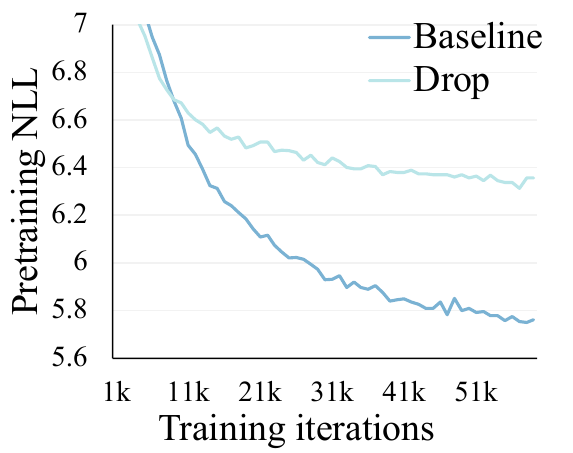}
        \caption{Drop tokens}
        \label{fig:ablation_drop}
    \end{subfigure}
    % \quad
     \begin{subfigure}[b]{0.24\linewidth}
        \includegraphics[width=0.99\linewidth]{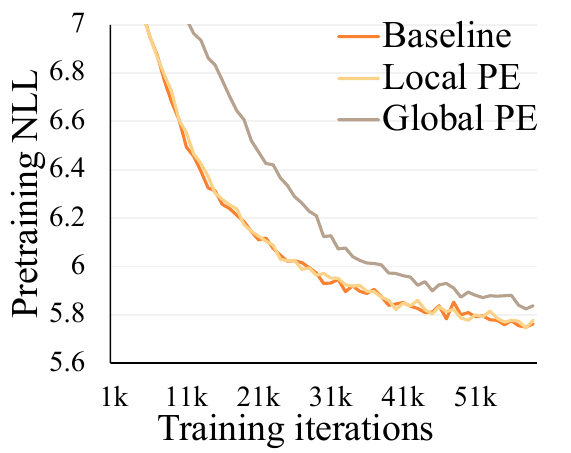}
        \caption{Positional Embedding}
        \label{fig:ablation_pe}
    \end{subfigure}
    % \quad
    \begin{subfigure}[b]{0.24\linewidth}
        \includegraphics[width=0.99\linewidth]{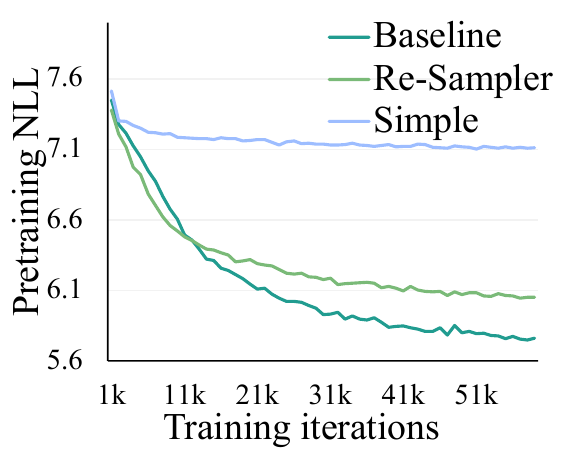}
        \caption{Re-sampler \& Simple impl.}
        \label{fig:ablation_resampler_simple}
    \end{subfigure}
    \vspace{-3mm}
\caption{
Effectiveness comparison of various Token-Shuffle implementations and alternatives. Our implementation shows reasonable alignment with the Token-Shuffle concept, as indicated by the training loss in a fair comparison.}
    \label{fig:albation_implementations}
    \vspace{-4mm}
\end{figure*}

For a fair comparison, we standardize all training configurations across these experiments. All models are trained for 60K iterations on 32 GPUs with a learning rate of $2 \times 10^{-4}$, a sequence length of 4096, and a batch size of 4. We conduct experiments at a resolution of $512 \times 512$, using a Token-Shuffle window size of 2 for all model variants. This setup allows us to directly compare training loss to evaluate the effectiveness of each design choice.

As shown in Fig.~\ref{fig:albation_implementations}, the training loss (log-scaled perplexity, which is commonly used evaluation for pretraining stage) suggests that our default configuration is a reasonable choice for implementing Token-Shuffle. In Fig.~\ref{fig:ablation_depth}, we observe that adding more MLP blocks in the Token-Shuffle operations (for both input and output) does not lead to noticeable improvements. Additionally, Fig.~\ref{fig:ablation_drop} illustrates that retaining all visual tokens is crucial.
Our experiments further reveal that additional positional embeddings do not enhance Token-Shuffle, likely because MLP layers are inherently position-aware and RoPE is already employed to model relative positional information among fused visual tokens. We also observe that the Re-sampler performs worse than our Token-Shuffle as demonstrated in Fig.~\ref{fig:ablation_resampler_simple}; this may be due to our Re-sampler’s design, which is forced for local fusion and disentanglement, differing from original Re-sampler in SEED-X and related works. Meanwhile, the simplified version of our method performs the worst, even though it introduces more parameters, possibly due to the linear projection and overly simplified output design — an area for further investigation.

\subsubsection{Comparison of different shuffle sizes}
\begin{wrapfigure}{r}{0.5\textwidth}
    \centering
    \includegraphics[width=0.8\linewidth]{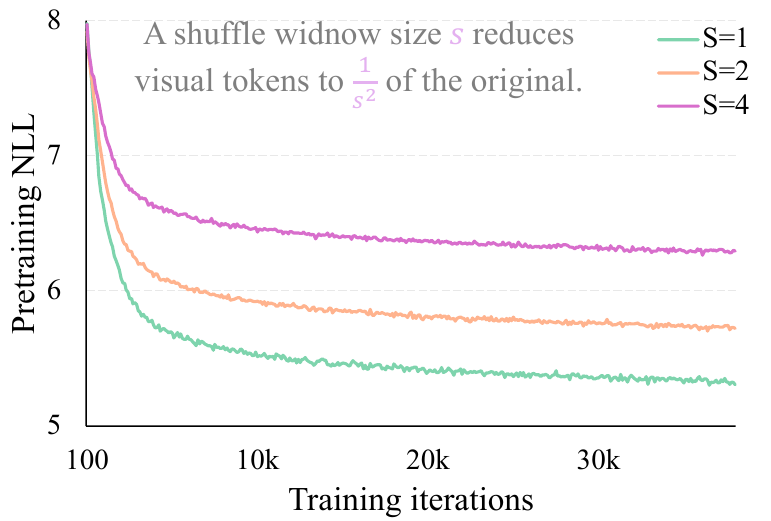}
    \vspace{-3mm}
    \caption{Training losses for different shuffle window sizes.}
    \label{fig:train_loss_s1s2s4}
    \vspace{-5mm}
\end{wrapfigure}
Our Token-shuffle enjoys flexible settings of Token-Shuffle window size, like 1, 2, 4, and even larger, resulting in different levels of token compression and efficiency boosts. However, we acknowledge that larger Token-Shuffle window size will certainly decrease generation quality due to significantly reduced computations in Transformer. Here, we investigate the impact of different shuffle window sizes in Fig.~\ref{fig:train_loss_s1s2s4}.

Note that a shuffle window size of 1 implies that no Token-Shuffle is applied, though additional MLP layers are still introduced. As expected, increasing the shuffle window size leads to higher training loss and a corresponding reduction in generation quality. This is a logical and anticipated phenomenon, as a single fused token represents an increasingly larger number of visual tokens and significant computational reduction for Transformer. Exploring methods to minimize the gap in quality and training loss remains an important area of interest.
Fig.~\ref{fig:visualization_s1s2s4} illustrates the differences in generated images across various shuffle sizes, with each image labeled with its VQAScore~\cite{lin2024evaluating}. When the shuffle size is small, such as 1 or 2, the generated images exhibit excellent quality. With larger shuffle sizes, while high-fidelity images are still achievable, a slight blurring effect is noticed. Extended training could potentially help mitigate this issue.

\section{Conclusion}
In this work, we introduce Token-Shuffle for efficient and scalable image generation in MLLMs. Unlike prior methods that rely on high compression ratios or reduced visual token inputs, we shuffle spatially local visual tokens for input and unshuffle the fused tokens back for output. Token-Shuffle is a lightweight, plug-and-play design for MLLMs that adheres to the next-token prediction paradigm while enabling batch generation of tokens within a local window. Our Token-Shuffle significantly reduces computational cost and accelerates inference. Leveraging these advantages, for the first time, we push the boundaries of autoregressive text-to-image generation to a resolution of $2048 \times 2048$, achieving high efficiency in training and inference at low cost while maintaining promising generation quality.  As a tentative exploration, we anticipate further advancements toward scalable image generation for autoregressive models.

\clearpage
\newpage
\bibliographystyle{assets/plainnat}
\bibliography{paper}

\clearpage
\newpage
\beginappendix

This supplementary material provides more implementation details, ablation studies, visualization results, discussions and limitations. We provide detailed implementations in Sec.~\ref{sec:implementation_details} to provide more insights. We also present more studies and visualization results in Sec.~\ref{sec:more_studies}. Finally, we discuss the limitations and further work of Token-Shuffle  in Sec.~\ref{sec:discussions}.
\section{Implemental Details}
\label{sec:implementation_details}

\begin{figure}[!h]
    \centering
    \includegraphics[width=0.4\linewidth]{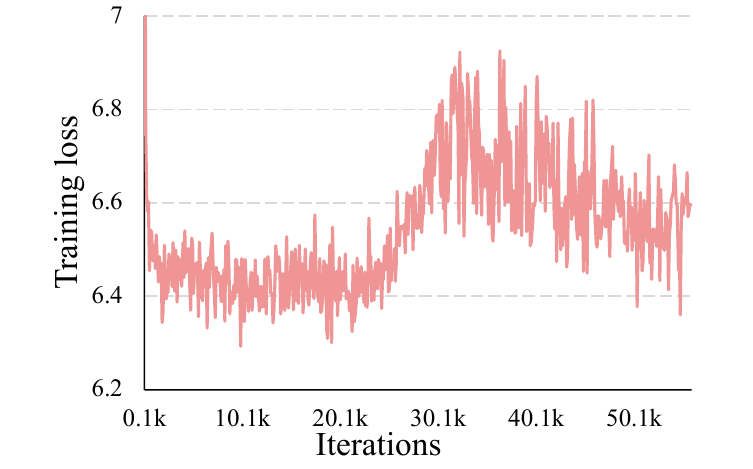}
    \hspace{5mm}
    \includegraphics[width=0.4\linewidth]{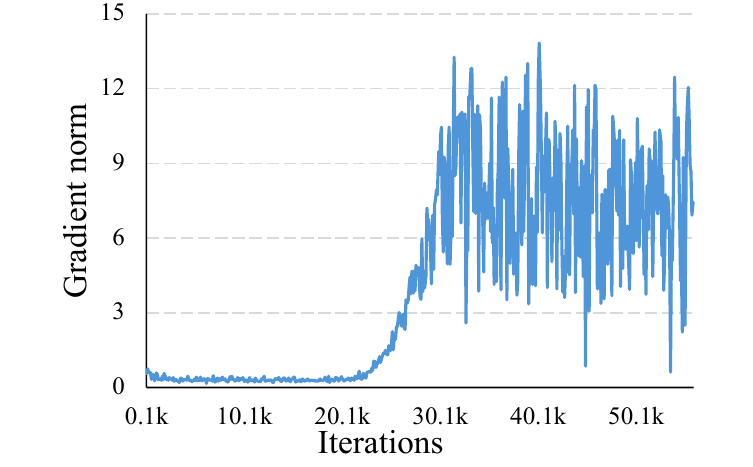}
    \caption{We plot the \textcolor{Salmon}{average loss (left)} and \textcolor{RoyalBlue}{gradient norm (right)} when training with a resolution of $2048\times 2048$. Training shows instability after approximately 20K iterations.}
    \label{fig:2048_unstable}
\end{figure}

\paragraph{Instability in training 2048 resolution}  Training at resolutions of $512\times512$ or $1024\times1024$ is notably stable, with the loss consistently decreasing throughout the process. However, \textit{training at very high resolutions, such as $2048\times2048$, often becomes unstable}, as evidenced by a significant increase in training loss after several thousand iterations, as illustrated in Fig.~\ref{fig:2048_unstable}.

To investigate the cause of unstable training, we analyze the training process in detail. Initially, we hypothesize that the instability arises from using a large learning rate, a common factor in such issues. To test this, we reduce the learning rate from $1e^{-4}$ to $5e^{-5}$ and $1e^{-5}$, decreasing it by factors of 2 and 10, respectively. However, the training instability persists, suggesting that the learning rate is not the root cause.
Next, inspired by EMU3~\cite{wang2024emu3}, we consider that high-resolution images might cause visual tokens to dominate the training process. To address this, we apply a loss weight of $0.5$ or $0.2$ to the visual tokens. Unfortunately, this adjustment also fails to stabilize the training.
We then investigate whether the logit shift issue, which has been observed to cause unstable training in larger models such as Chameleon~\cite{team2024chameleon} and Lumina-mGPT~\cite{liu2024lumina}, could also occur in our 2.7B model. Notably, this phenomenon is typically associated with models containing 7B parameters or more. To tackle this, we consider two solutions: (1) incorporating QK-Norm into each attention layer, and (2) adding z-loss~\cite{team2024chameleon} to the training objective.
Empirically, we find that while QK-Norm partially alleviates the issue, the instability eventually recurs as training progresses. In contrast, z-loss effectively prevents instability throughout training. Thus, we combine both QK-Norm and z-loss to stabilize the training at $2048\times2048$ resolution, and set the z-loss weight to $1e^{-5}$. Retrospectively, we emphasize that z-loss not only helps large models as indicated in Chameleon and Lumina-mGPT, but also helps very high-resolution image generation for discrete image generation pipeline. 

\paragraph{Inference Implementation}
We consider both textual tokens and visual tokens for loss backpropagation, which has been empirically proven to be beneficial for text faithfulness. This approach trains both text and images, aligning with the philosophy of MLLMs, with the key difference being that we only use text-image paired datasets. However, during inference, the model (as with all autoregressive models) may (1) continue generating text instead of an image, or (2) produce mixed text-image tokens, resulting in incomplete images.

\begin{figure*}[!h]
    \centering
    \includegraphics[width=1\linewidth]{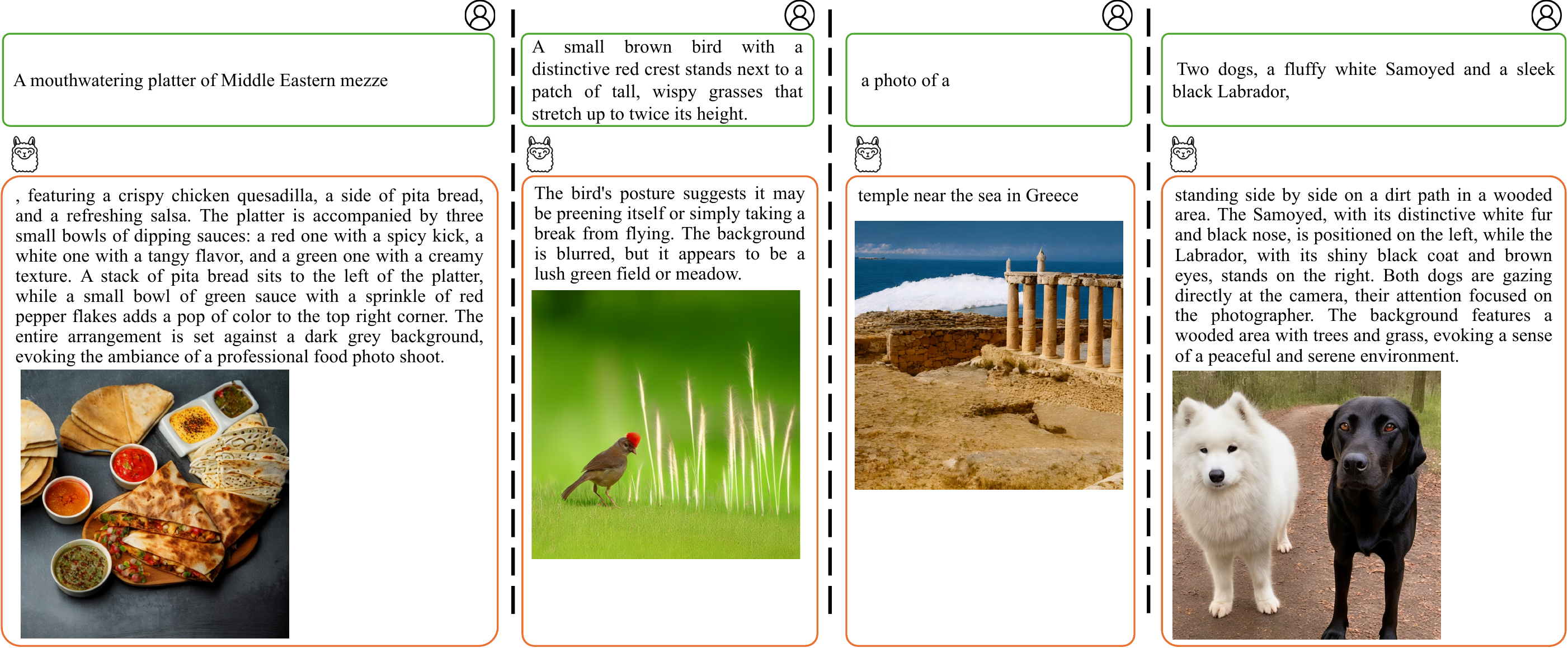}
    \vspace{-3mm}
    \caption{Without explicitly appending \texttt{<|start\_of\_image|>} token, our model naturally generates text based on input and seamlessly transitions to an image, consistently and automatically concluding in line with training data format.}
    \label{fig:generate_text_image}
    \vspace{-5mm}
\end{figure*}

To address these issues, we first introduce a special token, \texttt{<|start\_of\_image|>}, appended to the end of prompt tokens. This ensures that the model always generates an image after the prompt. Without this token, the model may generate additional text as a supplement to simple prompts before concluding with an image, as shown in Fig.~\ref{fig:generate_text_image}.

For mixed text-image tokens, we observe that during the early stages of training, the model is more prone to generating such outputs. However, as training progresses, the model consistently generates visual tokens up to the \texttt{<|end\_of\_image|>} token, resulting in complete images. In rare cases where this behavior does not occur, we enforce structural generation by restricting tokens following \texttt{<|start\_of\_image|>} to be sampled only from the visual vocabulary, ensuring the generation of complete images.

\begin{figure}[]
  \centering
  \begin{minipage}[c]{0.43\textwidth}
    \centering
    \includegraphics[width=1\linewidth]{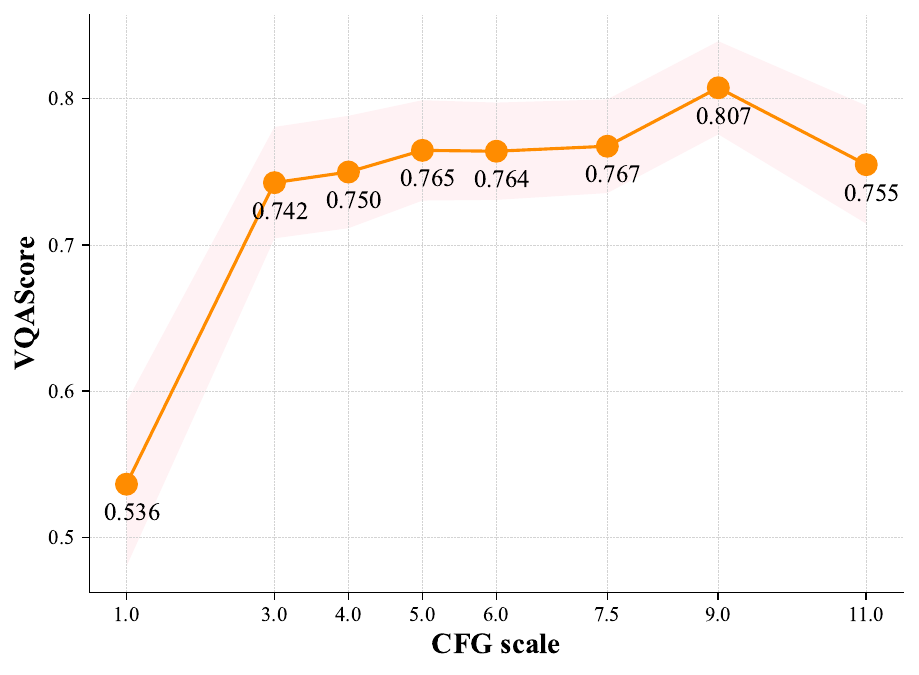}
    \vspace{-6mm}
    \caption{CFG scale \textit{vs.} VQAScore.}
    \label{fig:cfg_vqa}
  \end{minipage}
  \hfill
  \begin{minipage}[c]{0.5\textwidth}
    \centering
    \vspace{8mm}
    \includegraphics[width=1\linewidth]{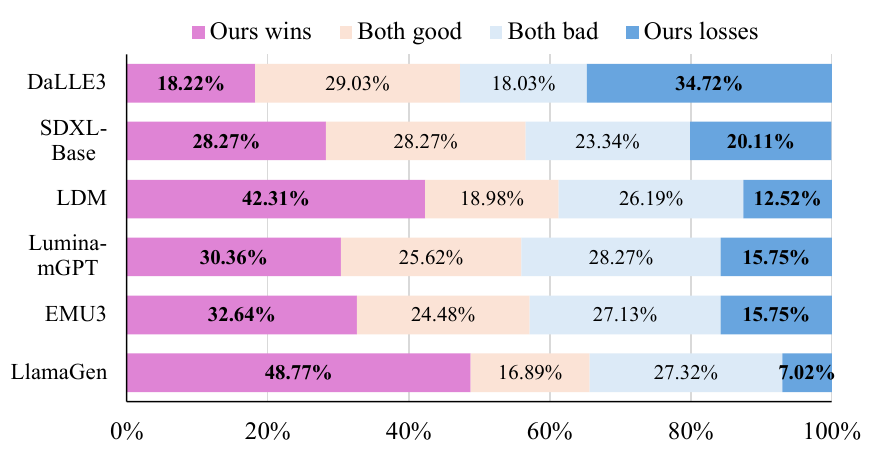}
    \vspace{-5mm}
    \caption{Human evaluation of text alignment, comparing Token-Shuffle with various AR-based and diffusion-based models. Results may vary slightly from Fig.~\ref{fig:human_eval} due to the generated images are assessed by different vendors.}
    \label{fig:supp_text_alignment}
  \end{minipage}
\end{figure}

\section{More Studies}
\label{sec:more_studies}

\subsection{Choice of CFG scales}

\begin{figure*}
    \centering
    \includegraphics[width=0.99\linewidth]{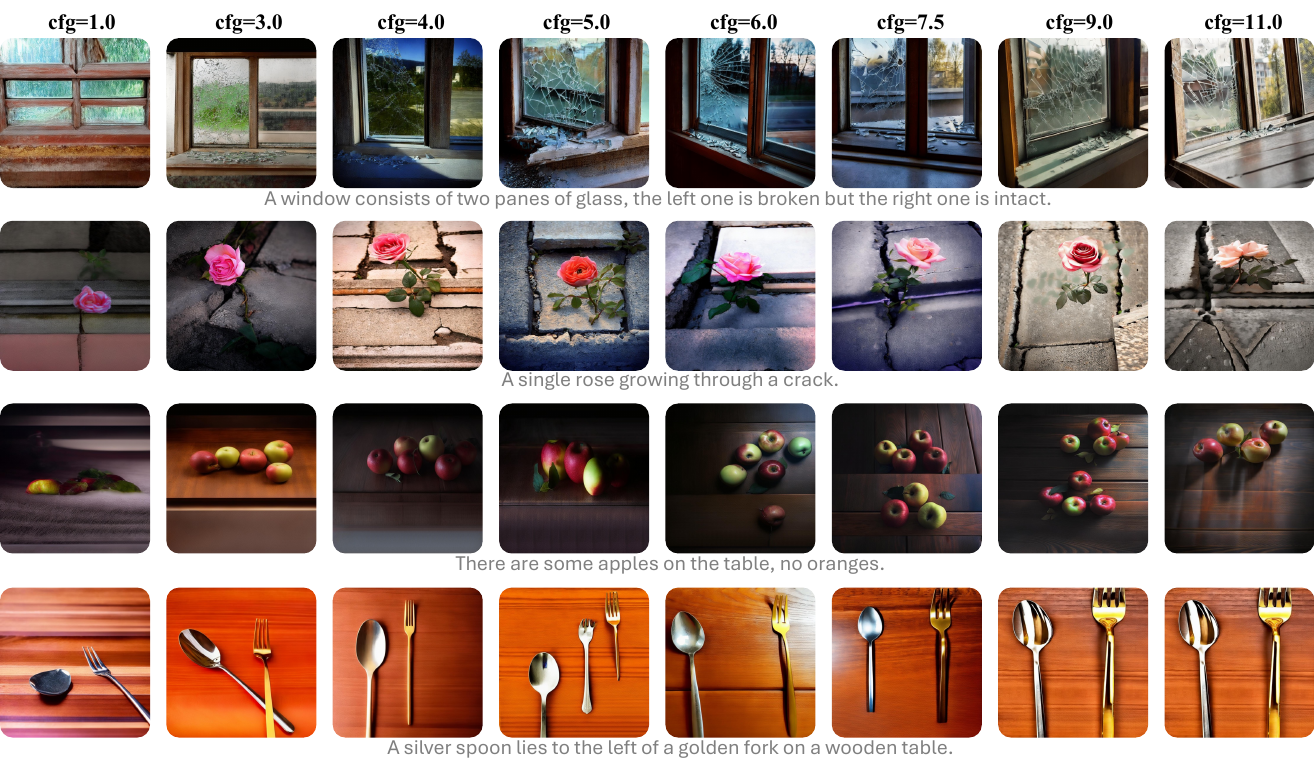}
    \vspace{-3mm}
    \caption{Examples of generated images under different CFG scales.}
    \label{fig:cfg_scales_examples}
\end{figure*}

Conceptually, CFG enhances generation quality by balancing prompt fidelity with visual coherence. However, determining the optimal CFG scale is empirical and model-dependent~\cite{girdhar2023emu, sun2024autoregressive, li2024autoregressive, peebles2023scalable, tian2024visual}. We systematically evaluate different CFG scales, ranging from 1.0 to 11.0, with VQAScore results presented in Fig.\ref{fig:cfg_vqa} and illustrative examples shown in Fig.\ref{fig:cfg_scales_examples}. It is worth noting that no CFG schedulers were introduced in this study.

While a higher CFG scale generally leads to improved VQAScore, as demonstrated in Fig.\ref{fig:cfg_vqa}, we observe that it may also result in a slight deterioration of visual appearance, as illustrated in Fig.\ref{fig:cfg_scales_examples}. Taking into account both the qualitative and quantitative findings presented, we consider that a CFG value of 7.5 strikes the optimal balance between performance and visual quality.

\subsection{Text Alignment}
We observe that our model delivers superior text-alignment performance, as demonstrated in the human evaluation results in Fig.\ref{fig:human_eval}. To further substantiate this, we provide a detailed comparison, evaluating our method against additional models, with the corresponding human evaluation results presented in Fig.\ref{fig:supp_text_alignment}. Our images are generated using a half-linear CFG scheduler with a scale of 7.5 and a fixed random seed.

Clearly, Token-Shuffle significantly outperforms all other methods by a considerable margin, except for DALL-E 3, which also trains and infers on long prompts. This experiment highlights the effectiveness of using long and detailed captions to improve text-to-image (T2I) text-faithfulness.

\subsection{Causal Attention Mask}
\begin{wrapfigure}{r}{0.55\textwidth}
    \centering
    \vspace{-8mm}
    \includegraphics[width=1\linewidth]{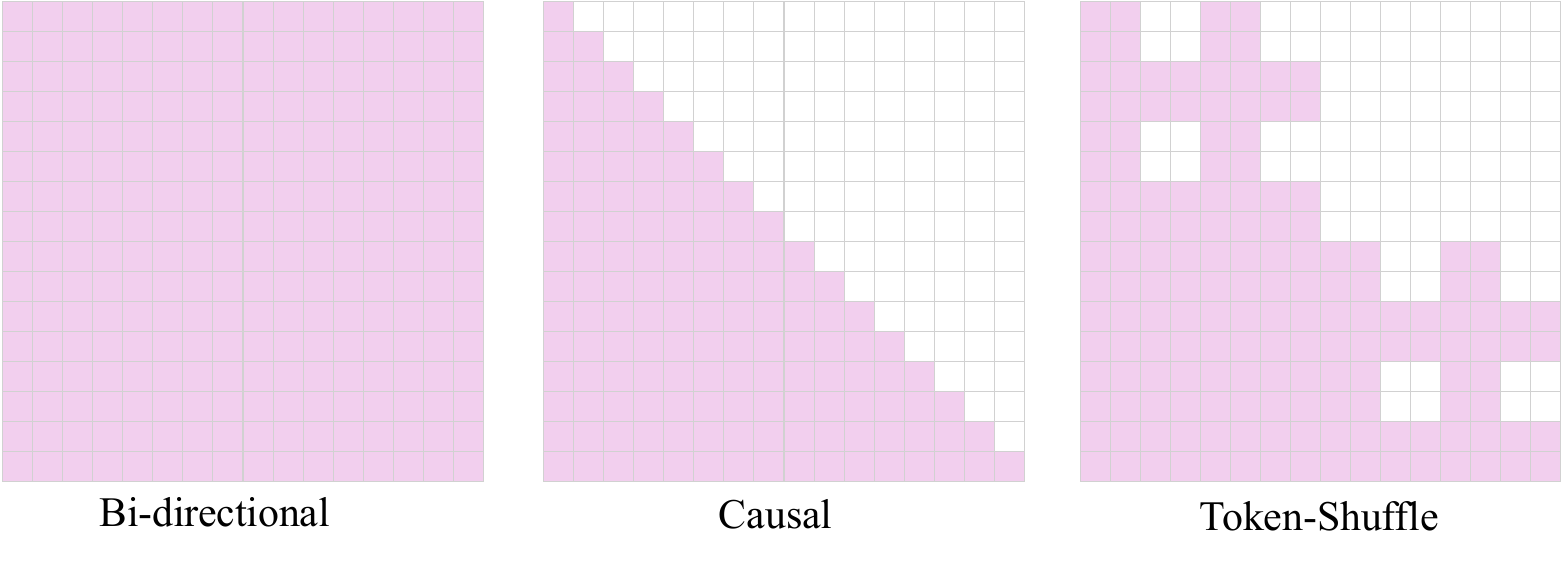}
    \vspace{-7mm}
    \caption{Attention maps of three implementations: bi-directional, causal, and Token-Shuffle. Illustrated with a feature map size of $4\times 4$ (16 tokens) and a shuffle window size of 2 for Token-Shuffle.}
    \label{fig:supp_causal_mask}
\end{wrapfigure}
\textit{Token-Shuffle adheres to the standard next-token prediction mechanism without altering the original causal mask used in LLMs.} However, instead of predicting the next single token, it predicts a fused token, which is then disentangled into spatially local tokens. In this approach, the fused token retains the same causal mask as the LLM, but the disentangled tokens introduce a modified causal mask that allows mutual interactions within the spatial local window. Fig.~\ref{fig:supp_causal_mask} compares the attention maps of bi-directional, causal, and Token-Shuffle implementations.

While the bi-directional implementation facilitates global token interactions and the causal implementation enforces strict sequential constraints, Token-Shuffle strikes a balance by enabling local mutual interactions among tokens. This design is anticipated to improve visual generation quality, particularly in capturing finer local details, compared to the traditional causal design. Please note that this is achieved without altering the causal masking for both training and inference.

\subsection{High-Compress VQGAN or Token-Shuffle}
\label{sec:compress_ratio}

\begin{figure}[]
  \centering
  \begin{minipage}[c]{0.55\textwidth}
    \centering
    \includegraphics[width=\textwidth]{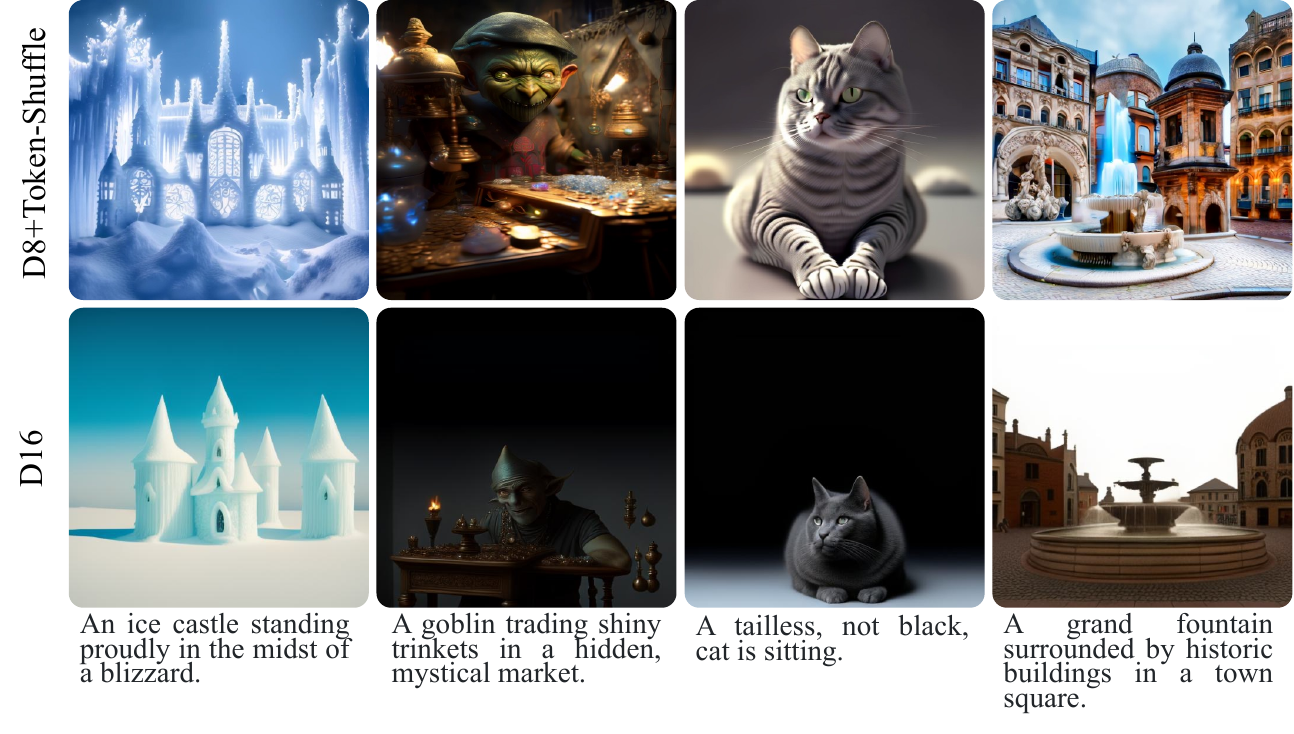}
    \caption{Visual examples comparing Token-Shuffle (compress ratio $8\times$ with Token-Shuffle window size of 2) and high compress VQGAN (compress ratio $16\times$).}
    \label{fig:down8_and_ts_EXP}
  \end{minipage}
  \hfill
  \begin{minipage}[c]{0.43\textwidth}
    \centering
    \vspace{8mm}
    \includegraphics[width=0.99\linewidth]{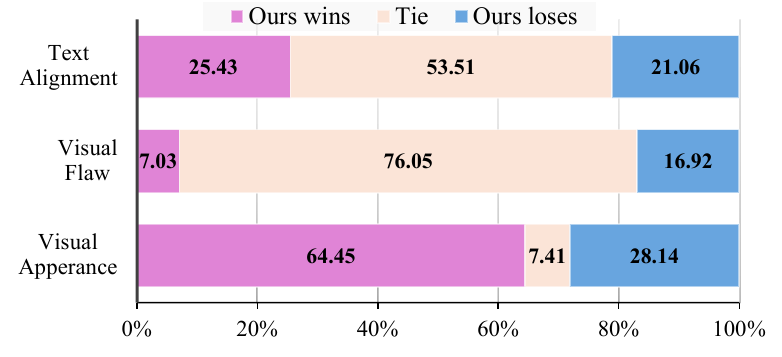}
    \vspace{7mm}
    \caption{Human evaluation of Token-Shuffle (compress ratio $8\times$ with Token-Shuffle window size of 2) and high compress VQGAN (compress ratio $16\times$).}
    \label{fig:down8_and_ts_EXP_human_eval}
    
  \end{minipage}
\end{figure}

Token-Shuffle incorporates additional lightweight layers into Transformers to reduce the number of tokens, enabling efficient processing and high-resolution image generation. In contrast, some concurrent efforts in the diffusion model field, such as SANA~\cite{xie2024sana}, adopt a high-compression VAE image tokenizer strategy (\textit{e.g.,} using a down-sampling ratio of $32\times$ rather than the more common $16\times$ or $8\times$). Here, we empirically explore and compare these two strategies (High-Compression Image Tokenizer \textit{vs}. Token-Shuffle) and then discuss their potential limitations.

\begin{wraptable}{r}{0.45\textwidth}
    \centering
    \resizebox{1\linewidth}{!}{
    \begin{tabular}{c|ccccc}
    \toprule
        Ratio&Tokens & Codebook & PSNR & SSIM &CLIP   \\
        \midrule
         Low ($8\times$)& 4096 & 8192 & 27.10 & 0.78 & 0.98 \\
         High ($16\times$)& 1024 & 16384 & 22.89 & 0.64 & 0.96 \\
     \bottomrule
    \end{tabular}
    }
    \caption{Reconstruction results of VQGAN models with different compress ratios. The results are achieved on MSCOCO-val set with a resolution of 512.}
    \label{tab:vqgan_reconstruction}
\end{wraptable}
For the comparison, we utilize two VQGAN models with different compression ratios: $16\times$ and $8\times$. The $16\times$ VQGAN model is taken from the previous LlamaGen T2I checkpoint, while the $8\times$ VQGAN is derived from our internal checkpoint. We first benchmark both models on the MSCOCO-val dataset~\cite{lin2014microsoft}, which consists of $5K$ images. The images are resized and center-cropped to a resolution of $512\times512$. The performance comparison of the VQGAN models is summarized in Tab.\ref{tab:vqgan_reconstruction}.

Clearly, a higher compression ratio significantly degrades reconstruction performance, which can negatively impact generation quality. Building on this observation, we investigate the generation quality of the two strategies using the aforementioned high- and low-compression VQGAN models. For this study, we generate $512\times 512$ resolution images, employing the $8\times$ compression ratio VQGAN with Token-Shuffle (shuffle window size of 2) to represent our Token-Shuffle strategy, and the $16\times$ compression ratio VQGAN to represent the high-compression image tokenizer approach. This setup ensures equivalent training and inference computational costs (excluding the negligible additional parameters and FLOPs introduced by Token-Shuffle).
All images are generated using the same settings, including identical CFG values, temperature, CFG scheduler, \textit{etc}. We evaluate and compare the two strategies on GenAI-Bench, reporting VQAScore and human evaluation results in Tab.~\ref{tab:down8_and_ts_EXP_vqascore} and Fig.~\ref{fig:down8_and_ts_EXP_human_eval}, respectively.

Both auto-evaluation and human evaluation results unequivocally demonstrate that Token-Shuffle consistently outperforms its high-compression VQGAN counterpart. For illustration, we also provide visual examples in Fig.~\ref{fig:down8_and_ts_EXP}.
However, we admit that this comparison is not entirely fair for the following reasons:
(1) The image tokenizers were not trained under identical conditions, and it is challenging to obtain fairly trained VQGAN models with different down-sampling ratios.
(2) During the course of our project, the dataset underwent slight and progressive changes—some images were added, while others were filtered out due to privacy concerns—affecting both pre-training and fine-tuning stages.
Despite these factors, we believe they do not impact the validity of our conclusions.

In general, a higher-compression VQGAN offers the simplest implementation for supporting efficient and high-resolution image generation; however, it compromises generation performance, as shown in Tab.\ref{tab:vqgan_reconstruction}, Tab.\ref{fig:down8_and_ts_EXP}, Fig.\ref{fig:down8_and_ts_EXP_human_eval}, and examples in Fig.\ref{fig:down8_and_ts_EXP}.
In contrast, Token-Shuffle, inspired by dimensional redundancy, introduces a pair of plug-and-play token operations that not only achieve superior generation performance and present better details but also provide dynamic settings for different shuffle window sizes, enabling adjustable compression results—a flexibility not available with high-compression VQGAN.

\begin{table*}
\centering
\resizebox{0.95\linewidth}{!}{
\begin{tabular}{lcccccccccccccc}
\toprule
\multirow{3}{*}{\textbf{Model}} &\multicolumn{6}{c}{\textbf{"Basic" prompts}}&\color{white}{.} & \multicolumn{6}{c}{\textbf{"Hard" prompts}}  \\
\cmidrule{2-7} \cmidrule{9-14}
&\multirow{2}{*}{\textbf{Attribute}}& \multirow{2}{*}{\textbf{Scene}} & \multicolumn{3}{c}{\textbf{Relation}} & \multirow{2}{*}{\textbf{Overall}} &  &\multirow{2}{*}{\textbf{Count}} & \multirow{2}{*}{\textbf{Differ}} & \multirow{2}{*}{\textbf{Compare}}& \multicolumn{2}{c}{\textbf{Logical}} & \multirow{2}{*}{\textbf{Overall}}\\
\cmidrule{4-6} \cmidrule{12-13}
 & & & Spatial & Action & Part & &  & & & & Negate & Universal &\\
\midrule
D16 & 0.80 & 0.82 & 0.79 & 0.79 & 0.86 & 0.80& & 0.72 & 0.71 & 0.73 & 0.65 & 0.75 & 0.71 \\
D8+TS&0.82 & 0.85 & 0.82 & 0.82 & 0.84 & 0.82 & & 0.77 & 0.77 & 0.77 & 0.66 & 0.74 & 0.72\\
\bottomrule
\end{tabular}
}
\vspace{-2mm}
\caption{\textbf{VQAScore evaluation of image generation on GenAI-Bench.} "D16" indicates directly using a high-compress VQGAN with a down-sampling ratio of $16\times$. "D8+TS" indicates using a low-compress VQGAN with a down-sampling ratio of $8\times$ and Token-Shuffle window size of 2.}
\label{tab:down8_and_ts_EXP_vqascore}
\end{table*}

\subsection{More visual examples}
\begin{figure*}
    \centering
    \includegraphics[width=0.99\linewidth]{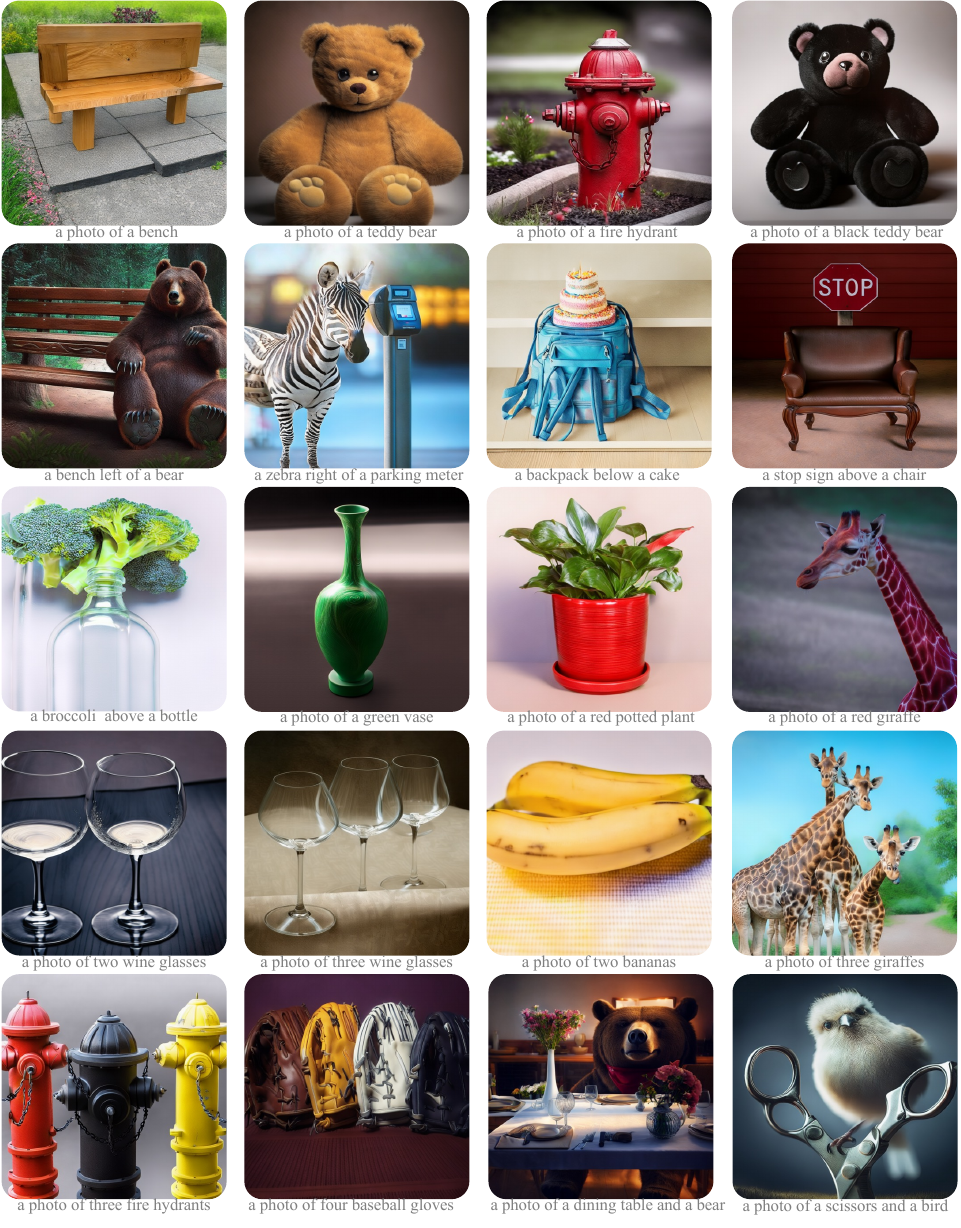}
    \caption{$\mathbf{1024\times 1024}$ resolution images generated by Token-Shuffle with a shuffle window size of 2. We show generated images focusing on position, color, counting, and combination. The prompts are from GenEval~\cite{ghosh2024geneval} prompts.}
    \label{fig:supp_1024_simple}
\end{figure*}

\begin{figure*}
    \centering
    \includegraphics[width=0.98\linewidth]{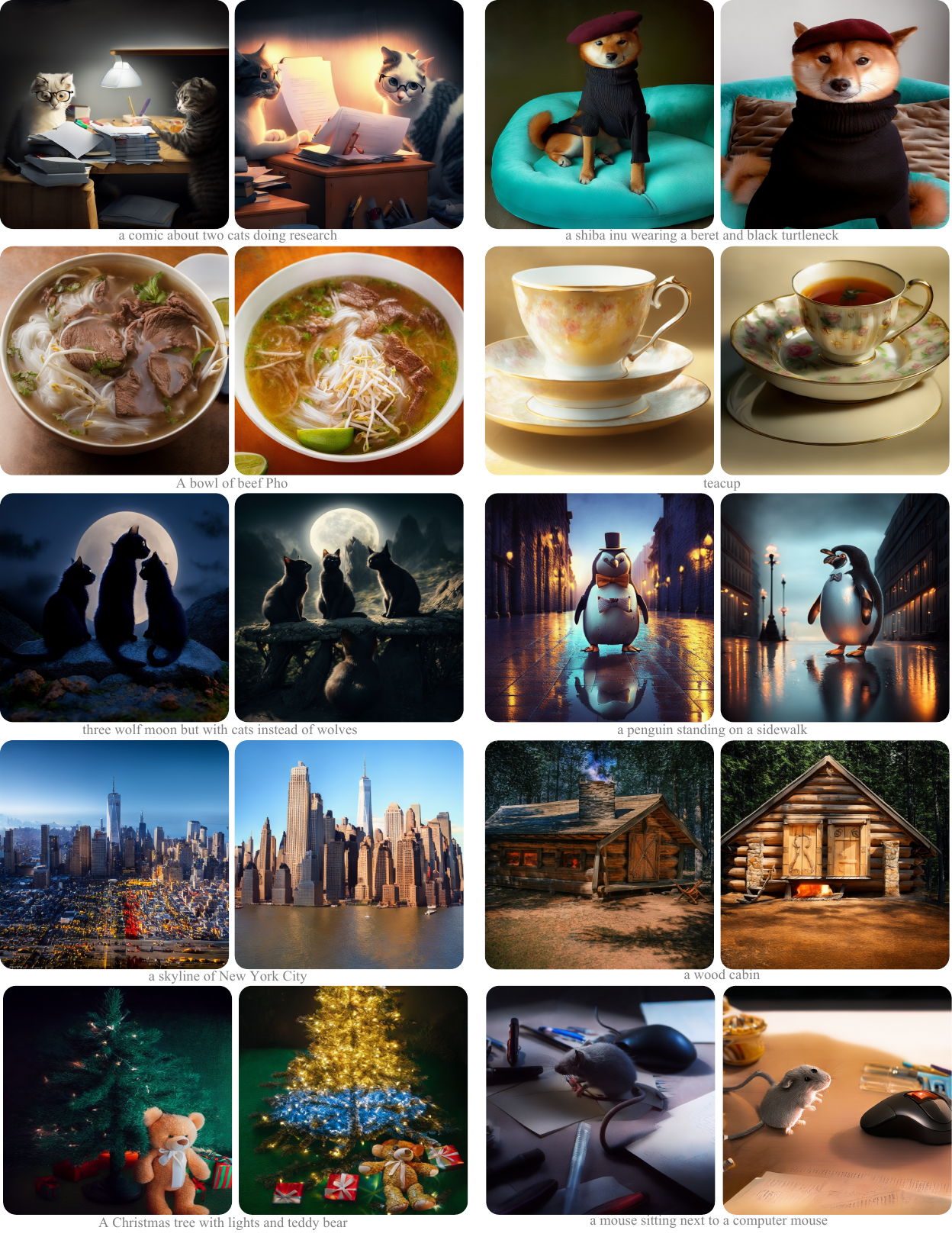}
    \vspace{-5mm}
    \caption{$\mathbf{1024\times 1024}$ resolution images generated by Token-Shuffle with a shuffle window size of 2. We show two images of same prompt with different random seeds, focusing on complex scenarios or hard prompts. The prompts are from our internal evaluation prompts.}
    \label{fig:supp_1024_ours}
\end{figure*}

\begin{figure*}
    \centering
    \includegraphics[width=1\linewidth]{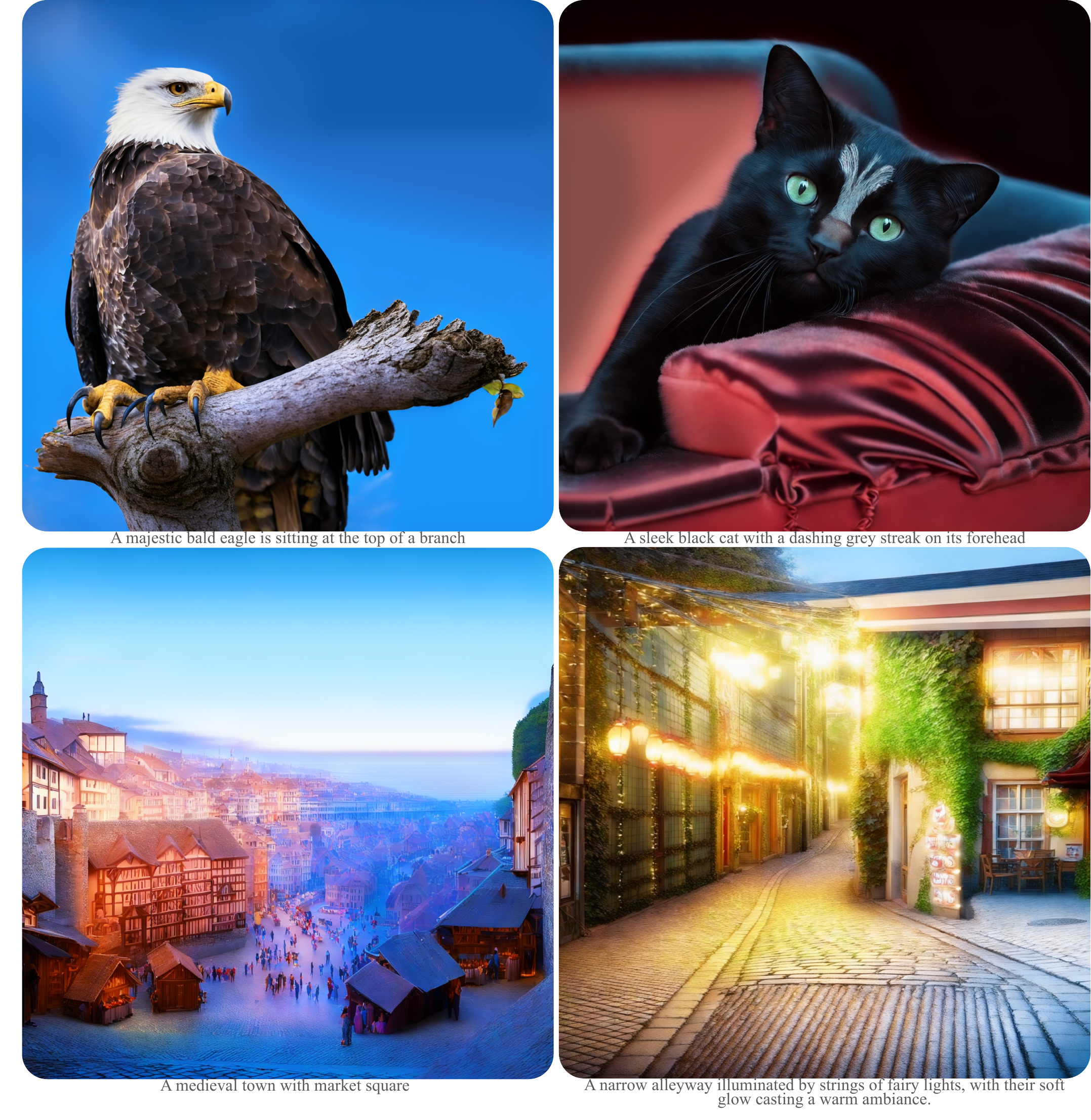}
    \vspace{-5mm}
    \caption{$\mathbf{2048\times 2048}$ resolution images generated by Token-Shuffle with a shuffle window size of 2. Images are resized for visualization. Please zoom in to see the details in top row and the overall soft holistic beauty in bottom row.}
    \label{fig:supp_2048}
    \vspace{-5mm}
\end{figure*}

We present additional visual examples in Fig.\ref{fig:supp_1024_simple} and Fig.\ref{fig:supp_1024_ours} to showcase the quality of $1024\times1024$ generated images. Further examples of $2048\times2048$ images are provided in Fig.~\ref{fig:supp_2048}. To our best knowledge, this is the first time AR-based models can generate such a high-resolution image efficiently and effectively. All images were generated with a shuffle window size of 2, half-linear CFG-scheduler with a scale of 7.5, as stated previously.

\section{Discussions}
\label{sec:discussions}
\subsection{Visual Flaws of AR-based models}
\begin{wraptable}{r}{0.6\textwidth}
    \centering
    \vspace{-5mm}
    \includegraphics[width=0.99\linewidth]{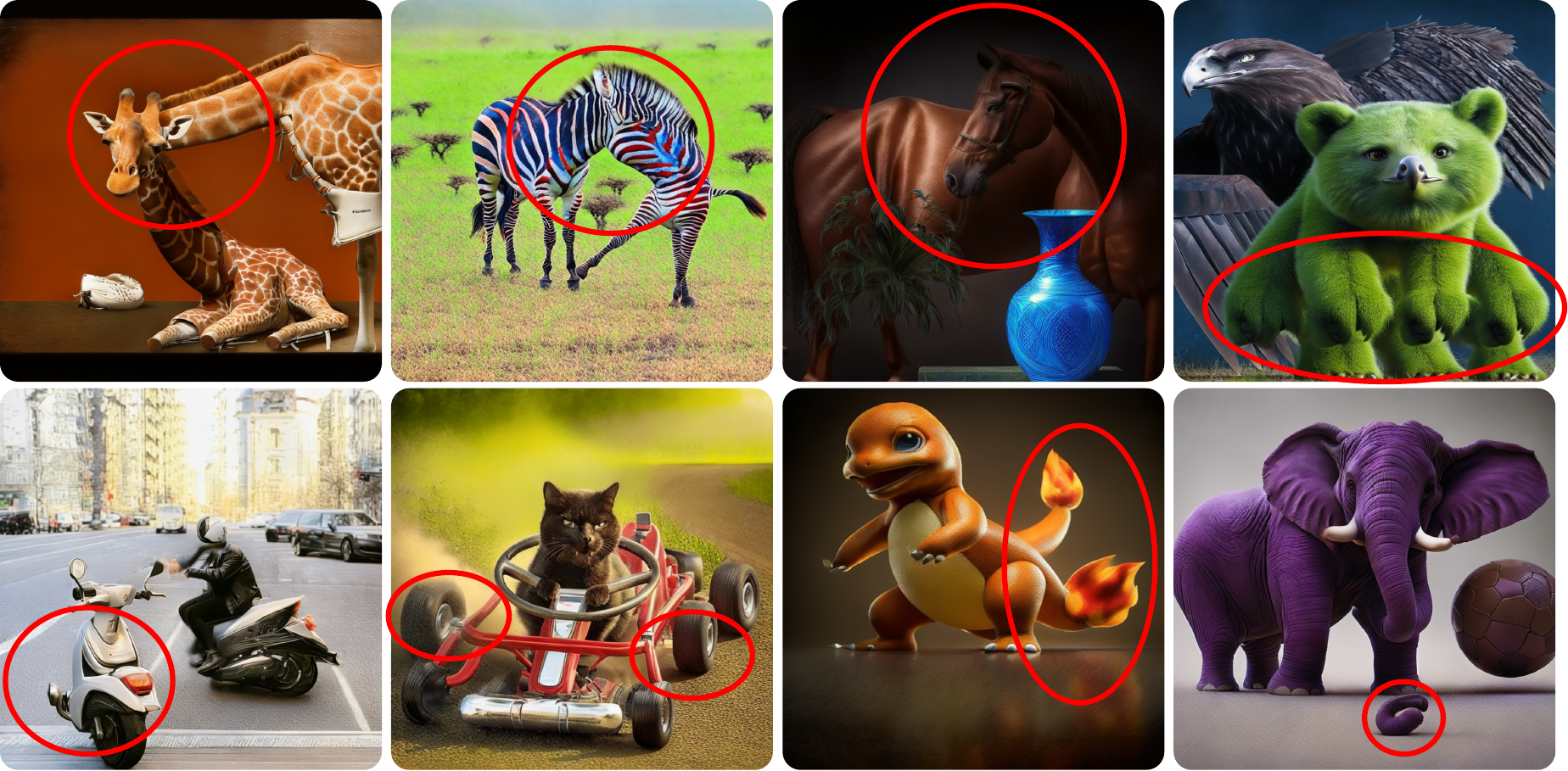}
    \vspace{-5mm}
    \caption{Examples of generated images with visual flaws and structural errors, marked with red circle (zoom in to see details).}
    \label{fig:failure_cases}
\end{wraptable}
As discussed in Fluid~\cite{fan2024fluid}, AR-based models often produce images with visual flaws (see the human evaluation comparison with LDM in Fig.\ref{fig:human_eval} (c)). This issue stems not from the information loss in VQGAN but from the limited global interaction inherent to causal masking and the next-token prediction framework. Although Token-Shuffle introduces local mutual interactions, it still struggles with this fundamental limitation. Fig.\ref{fig:failure_cases} shows examples of generated images with such visual flaws. Exploring approaches that maintain the next-token prediction framework while enabling global interactions remains an important direction for future research, with RAR~\cite{yu2024randomized} offering a promising starting point.
\subsection{Limitations} 
We introduce Token-Shuffle, targeting efficient high-resolution image generation with AR models with high quality. However, there are still interesting directions worth exploring. Firstly, we would like to see the scaling ability of Token-Shuffle in large LLMs, ~\textit{i.e.}, 7B and 30B models. We demonstrate that our 2.7B model is able to provide promising performance, outperforming 7B Lumina-mGPT, and can generate higher resolution. We expect better results when increasing the model size. Another interesting direction is to support  flexible resolutions, aspect ratios like EMU3~\cite{wang2024emu3}.

\end{document}